\documentclass{ieeeaccess}
\usepackage{cite}
\usepackage{amsmath,amssymb,amsfonts}
\usepackage{algorithmic}
\usepackage{graphicx}
\usepackage{textcomp}
\usepackage{url}
\usepackage{hyperref}
\usepackage{bm}
\usepackage{booktabs}
\usepackage{makecell}
\usepackage{placeins}
\usepackage{float}

\makeatletter
\AtBeginDocument{\DeclareMathVersion{bold}
\SetSymbolFont{operators}{bold}{T1}{times}{b}{n}
\SetSymbolFont{NewLetters}{bold}{T1}{times}{b}{it}
\SetMathAlphabet{\mathrm}{bold}{T1}{times}{b}{n}
\SetMathAlphabet{\mathit}{bold}{T1}{times}{b}{it}
\SetMathAlphabet{\mathbf}{bold}{T1}{times}{b}{n}
\SetMathAlphabet{\mathtt}{bold}{OT1}{pcr}{b}{n}
\SetSymbolFont{symbols}{bold}{OMS}{cmsy}{b}{n}
\renewcommand\boldmath{\@nomath\boldmath\mathversion{bold}}}
\makeatother

\def\BibTeX{{\rm B\kern-.05em{\sc i\kern-.025em b}\kern-.08em
    T\kern-.1667em\lower.7ex\hbox{E}\kern-.125emX}}

\begin{document}
\history{Date of publication xxxx 00, 0000, date of current version xxxx 00, 0000.}
\doi{10.1109/ACCESS.2024.0429000}

\title{Evaluating the Impact of Data Anonymization on Image Retrieval}
\author{\uppercase{Marvin Chen}\authorrefmark{1}, \uppercase{Manuel Eberhardinger}\authorrefmark{1}, \uppercase{and Johannes Maucher}\authorrefmark{1}}
\address[1]{Institute of Applied AI, Stuttgart Media University, Nobelstr. 10, 70569 Stuttgart, Germany}
\corresp{Corresponding author: Manuel Eberhardinger (e-mail: eberhardinger@hdm-stuttgart.de).}

\begin{abstract}With the growing importance of privacy regulations such as the General Data Protection Regulation, anonymizing visual data is becoming increasingly relevant across institutions. However, anonymization can negatively affect the performance of Computer Vision systems that rely on visual features, such as Content-Based Image Retrieval (CBIR). Despite this, the impact of anonymization on CBIR has not been systematically studied. This work addresses this gap, motivated by the DOKIQ project, an artificial intelligence-based system for document verification actively used by the State Criminal Police Office Baden-Württemberg. We propose a simple evaluation framework: retrieval results after anonymization should match those obtained before anonymization as closely as possible. To this end, we systematically assess the impact of anonymization using two public datasets and the internal DOKIQ dataset. Our experiments span three anonymization methods, four anonymization degrees, and four training strategies, all based on the state of the art backbone Self-Distillation with No Labels (DINO)v2. Our results reveal a pronounced retrieval bias in favor of models trained on original data, which produce the most similar retrievals after anonymization. The findings of this paper offer practical insights for developing privacy-compliant CBIR systems while preserving performance.
\end{abstract}


\titlepgskip=-15pt

\maketitle

\section{Introduction}
\label{sec:introduction}
Image retrieval systems have gained increasing importance in domains such as document analysis, security, and forensic investigations, enabling content-based search of visually similar images and thereby improving decision-making processes \cite{datta2008image, kokare2010document, chen2005content,wu2022crime, liu2024survey}. One such system is the retrieval component of the document verification system used in the DOKIQ research project\cite{dokiq}, which supports investigators from the State Criminal Police Office Baden-Württemberg (LKA BW) by retrieving visually similar cases from a database after scanning a new document. Despite the practical usefulness of image retrieval systems, ensuring compliance with data protection regulations such as the General Data Protection Regulation (GDPR) introduces additional challenges. Specifically, personal information, including faces and text, must be anonymized before storage and retrieval.

However, anonymization alters visual features in ways that may degrade retrieval quality and limit the utility of such systems. Despite its importance, the effect of anonymization on Content-Based Image Retrieval (CBIR) remains underexplored. Existing studies have focused mainly on how anonymization affects dense prediction tasks in computer vision, e.g., instance or semantic segmentation \cite{zhou2022impacts, hukkelaas2023does, lee2024balancing, stenger2024evaluating}, but its influence on retrieval performance has not been systematically assessed.

This work addresses this gap by evaluating the impact of anonymization on image retrieval performance and investigating whether training on anonymized data can mitigate performance degradation. We use DINOv2 \cite{oquab2023dinov2} as a backbone model as it can be seen as a state of the art model in visual representation learning and is also used in the DOKIQ retrieval component. We apply several anonymization techniques to two public datasets and the DOKIQ dataset to simulate privacy-compliant retrieval scenarios.

Our contributions are as follows:
\begin{itemize}
\item We systematically evaluate how various anonymization techniques affect image retrieval quality.
\item We examine whether training DINOv2 on anonymized data reduces performance loss, compared to DINOv2 trained on original data.
\item We propose metrics and methods for assessing anonymization-induced losses in retrieval quality.
\item We validate these metrics using downstream classification tasks.
\item We provide practical recommendations for the development of privacy-compliant CBIR-systems with the least possible loss of retrieval quality.
\end{itemize}

This paper is organized as follows. Section~\ref{sec:rel} reviews related work on the impact of data anonymization on computer vision performance and situates this study within the broader research landscape. Additionally, the DOKIQ project is introduced in more detail as a representative case of datasets containing documents with Personally Identifiable Information (PII). Section~\ref{sec:met} describes the methodological framework and experimental setup employed to evaluate the impact of anonymization on image retrieval. Section~\ref{sec:res} presents quantitative and qualitative results in both tabular and graphical form. Section~\ref{sec:dis} discusses the findings, formulates practical recommendations, and outlines potential directions for future research.

\section{Related Work}
\label{sec:rel}

\subsection{Data Anonymization}
\label{subsec:data-anomymization}
Data anonymization is a critical step for privacy-preserving computer vision applications. It aims to protect PII by modifying or removing sensitive content from visual data. Common approaches include traditional techniques such as blurring, pixelation, distortion, and masking \cite{lee2024balancing}. More recent research explores synthetic anonymization methods that replace PII, most notably human faces and bodies, with artificially generated content using generative models such as Generative Adversarial Network (GAN)-based \cite{hukkelaas2023deepprivacy2} or diffusion-based \cite{kung2024face} models.

\subsection{Impact of Data Anonymization on Computer Vision Performance}
\label{subsec:imp}
Stenger et al. \cite{stenger2024evaluating} summarized existing studies that perform systematic evaluations across multiple anonymization methods and computer vision tasks. We adopt their summary in Tab.~\ref{tab:anonymization_methods_comparison} and extend it by introducing two additional dimensions: the type of anonymized entities and the validation dataset used for performance assessment.

Zhou et al. \cite{zhou2022impacts} evaluated the impact of four anonymization techniques, Gaussian blurring, crop out, random crop, and GAN-based synthetic anonymization, on semantic segmentation performance using the Cityscapes dataset \cite{cordts2016cityscapes}. The study focused on anonymizing faces and license plates and tested multiple models using Residual Neural Network (ResNet) \cite{he2016deep} backbones. Their results show that all methods, except for image resynthesis, significantly degrade performance for human-related classes (e.g., pedestrians), while vehicle detection remains largely unaffected. Crop-based methods led to the strongest performance loss, followed by blurring. Image resynthesis achieved the best trade-off between privacy and utility. The authors also found that smaller models and higher-resolution inputs were more sensitive to anonymization effects.

Hukkel{\aa}s et al.~\cite{hukkelaas2023does} investigated the effect of various anonymization techniques, including traditional blurring and masking, as well as synthetic anonymization using Deep Privacy 2 (DP2) \cite{hukkelaas2023deepprivacy2}, on instance segmentation and pose estimation. They evaluated models on Cityscapes, BDD100K \cite{yu2020bdd100k}, and Common Objects in Context (COCO) \cite{lin2014microsoft} using models with ResNet backbones. Their results indicate that face anonymization had minimal impact on datasets where faces occupy little image area, but significantly degraded performance on COCO due to larger and more frequent facial regions. Synthetic anonymization consistently outperformed traditional methods. For full-body anonymization, all methods caused severe performance drops, with synthetic approaches again showing reduced degradation. Interestingly, contrary to Zhou et al., larger models (e.g., ResNet-152 compared to ResNet-50) exhibited higher sensitivity to anonymized training data on COCO, suggesting that generalization effects depend on dataset characteristics.

Lee et al. \cite{lee2024balancing} analyzed the effect of traditional anonymization methods, blurring, mask-out, pixelation, and distortion, on instance segmentation, pose estimation, and object detection using the COCO dataset. They were the first to evaluate You Only Look Once (YOLO) \cite{redmon2016you} and Real-Time Detection Transformer (RT-DETR) \cite{zhao2024detrs} models under anonymization. Distortion and mask-out led to the highest performance degradation. In line with Zhou et al., models with more parameters showed greater robustness. Additionally, Convolutional Neural Network (CNN)-based \cite{o2015introduction} architectures proved more resilient to anonymization than Vision Transformer (ViT)-based \cite{dosovitskiy2020image} ones, despite similar parameter counts. The study also noted that objects frequently cooccurring with humans experienced amplified performance loss when the humans in the image are anonymized.

Most recently, Stenger et al. \cite{stenger2024evaluating} evaluated the impact of various anonymization methods on face detection, keypoint detection, and instance segmentation using the COCO dataset. In contrast to previous studies, they assessed performance on anonymized validation data without retraining on an anonymized dataset, using only pre-trained models with ResNet backbones. Anonymization techniques included multiple variants of traditional methods (e.g., Gaussian, uniform, median blur, pixelation, mask-out), as well as synthetic methods (DP2, L-Diff \cite{kung2024face}). They also varied anonymization strength and region (face-mask, bounding box, entire image). L-Diff showed the best overall performance, while DP2 and traditional methods caused significant degradation, especially for instance segmentation and keypoint detection. Additionally, they simulated a re-identification attack using a subet of the Labeled Faces in the Wild (LFW) dataset \cite{huang2008labeled} to assess privacy risks. Mask-out on the face-mask region offered the strongest privacy, while DP2 had a 22 times higher re-identification risk than random retrieval, despite outperforming L-Diff in privacy terms.

\begin{table*}[t]
    \centering
    \caption[Comparison of previous work and this thesis evaluating the impact of anonymizing data on computer vision performance]{Comparison of previous work and this paper systematically evaluating the impact of data anonymization on computer vision performance. Adopted from Stenger et al. \cite{stenger2024evaluating} and extended by two additional columns.}
    \label{tab:anonymization_methods_comparison}
    \begin{tabular}{@{}p{1.5cm}|p{2.8cm}|p{2.3cm}|p{1.5cm}|p{2.0cm}|p{3cm}@{}}
        \toprule
        \textbf{Source} & \textbf{Anonymization methods} & \textbf{Anonymized entities} & \textbf{Retraining on anonymized data} & \textbf{Validation dataset} & \textbf{Vision tasks} \\
        \midrule
        \cite{zhou2022impacts}, 2022 & Synthetic, \newline traditional & Humans, \newline license plates & Yes & Original & Semantic segmentation \\
        \hline
        \cite{hukkelaas2023does}, 2023 & Synthetic, \newline traditional & Humans & Yes & Original & Instance/\newline semantic segmentation,\newline pose estimation \\
        \hline
        \cite{lee2024balancing}, 2024 & Traditional & Humans,\newline vehicles & Yes & Original & Instance/\newline semantic segmentation,\newline pose estimation,\newline object detection  \\
        \hline
        \cite{stenger2024evaluating}, 2024 & Synthetic, \newline traditional & Humans & No & Anonymized & Instance/\newline semantic segmentation,\newline keypoint detection,\newline face detection  \\
        \hline
        Ours, 2025 & Traditional & Humans,\newline documents & Yes & Anonymized & Image retrieval/\newline classification \\
        \bottomrule
    \end{tabular}
\end{table*}

\subsection{DOKIQ}
\label{subsec:dok}
The DOKIQ project \cite{dokiq}, a collaboration between Stuttgart Media University (HdM) and the LKA BW, serves as a practical motivation for this work. DOKIQ aims to support the forensic examination of documents,  e.g., identity documents, driver licenses or birth certificates, to name a few, through AI-based methods and includes a CBIR system that allows investigators to query both new and closed cases against previously closed cases. The CBIR component is built on top of DINOv2 and thereby leverages self-supervised learning to generate high quality embeddings from scanned documents, enabling retrieval without manual annotation.

To comply with the GDPR, DOKIQ is integrating an anonymization module that automatically redacts PII from scanned identity documents. This module builds on prior work by Eberhardinger et al. \cite{eberhardinger2025anonymization}, who proposed a retrieval-based anonymization pipeline using object detection and affine transformations. The pipeline minimizes information loss while ensuring legal compliance.

All closed cases in the DOKIQ system must undergo anonymization before being stored in the database, in accordance with the GDPR. DOKIQ's CBIR component can thus be viewed as a representative use case for similar systems that maintain databases containing documents with PII and are required to anonymize them prior to storage and processing. Understanding the impact of anonymization on tasks such as image retrieval is, therefore, of broad practical relevance.

\section{Methods}
\label{sec:met}
\subsection{General Idea}
\label{subsec:gen}
The goal of this work is to preserve the retrieval behavior of the original system after anonymization, enabling users to continue relying on the system despite privacy constraints. Therefore, we use the top 5\% of the retrieval results of a baseline model trained on the original, unanonymized data as a pseudo ground truth. We then compare the retrieval results of systems that work with anonymized data against this pseudo ground truth. The closer the post-anonymization results resemble the original retrievals, the more effective the anonymization approach is considered in preserving task utility. An overview of the approach is illustrated in Fig.~\ref{fig:experimental_setup}, which shows the pipeline from original training data to retrieval comparison after anonymization.

\Figure[th](topskip=0pt, botskip=0pt, midskip=0pt)[width=0.75\textwidth]{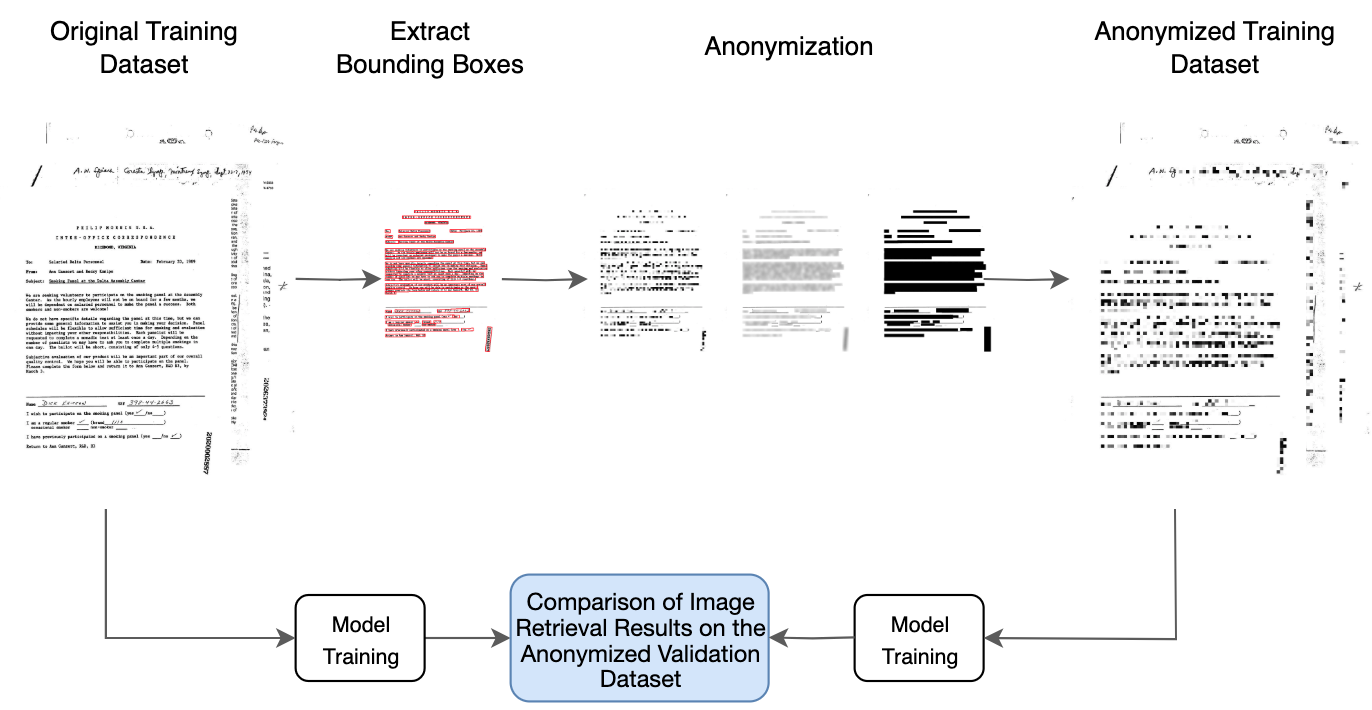}
{\textbf{Overview of the experimental setup. The model trained on the original training dataset provides pseudo ground truth retrievals, which serve as the evaluation baseline for both models adapted to anonymized data and the original model itself when applied to anonymized inputs.}\label{fig:experimental_setup}}

\subsection{Metrics}
\label{subsec:met}
To evaluate the performance of the anonymized retrieval systems, we employ two retrieval metrics: Mean Average Precision (mAP) and Mean Normalized Discounted Cumulative Gain (mnDCG). These metrics will be explained in the following.

\subsubsection{Average Precision (AP) and mAP}
Average Precision (AP) \cite{zhu2004recall} is a standard metric in image retrieval that combines precision and recall across different cut-off ranks. It is defined as:
\begin{equation}
    \mathrm{AP} = \sum_{k=1}^{n} P(k)\Delta r(k),
\end{equation}
where $P(k)$ is the precision at rank $k$ and $\Delta r(k)$ the change in recall from rank $k - 1$ to $k$. The definitions of precision (P) and recall (r) are as follows:
\begin{equation}
    P = \frac{|\{\text{relevant documents}\} \cap \{\text{retrieved documents}\}|}{|\{\text{retrieved documents}\}|},
\end{equation}
\begin{equation}
    r = \frac{|\{\text{relevant documents}\} \cap \{\text{retrieved documents}\}|}{|\{\text{relevant documents}\}|}.
\end{equation}
In this paper, the set of \{\text{relevant documents}\} is defined by the top 5\% of the original retrievals for each query image ordered in descending order based on cosine similarity. The overall performance across all query images is then aggregated using mAP:
\begin{equation}
    \mathrm{mAP} = \frac{1}{Q} \sum_{q=1}^{Q} \mathrm{AP}(q),
\end{equation}
where $Q$ denotes the total number of queries, which in our case corresponds to the number of validation images.

\subsubsection{Discounted Cumulative Gain (DCG) and mnDCG}
To also take into account the cosine similarities of the original system, we additionally use the Discounted Cumulative Gain (DCG) \cite{jarvelin2002cumulated}:
\begin{equation}
    \mathrm{DCG}_p = \sum_{i=1}^{p} \frac{\mathrm{rel}_i}{\log_2(i + 1)},
\end{equation}
where $\mathrm{rel}_i$ is in our case the cosine similarity-based relevance score, taken from the original retrievals, of the retrieved document at rank $i$ and $p$ denotes the top 5\% of the retrieved documents. We normalize the DCG by the Ideal Discounted Cumulative Gain (IDCG), which represents the DCG of a perfectly ordered result list, i.e., in our case the retrieval results of the baseline model:
\begin{equation}
    \text{IDCG}_p = \sum_{i=1}^{\lvert \text{REL}_p \rvert} \frac{2^{\text{rel}_i} - 1}{\log_2(i + 1)},
\end{equation}
where $|\mathrm{REL}_p|$ denotes the list of the original retrievals up to position $p$, sorted in descending order by cosine similarity.
The Normalized Discounted Cumulative Gain (nDCG) is then computed as:
\begin{equation}
    \mathrm{nDCG}_p = \frac{\mathrm{DCG}_p}{\mathrm{IDCG}_p}.
\end{equation}
The overall performance across all query images is then aggregated using mnDCG:
\begin{equation}
    \mathrm{mnDCG}_p = \frac{1}{Q} \sum_{q=1}^{Q} \mathrm{nDCG}_p(q).
\end{equation}

\subsection{Datasets and Anonymization}
\label{subsec:dat}
We selected two public datasets, Large-scale CelebFaces Attributes (CelebA) \cite{liu2015faceattributes} and Ryerson Vision Lab Complex Document Information Processing (RVL-CDIP) \cite{harley2015evaluation}, due to their proximity to the document-centric and identity-related domains relevant to our use case.

For the CelebA dataset, we applied the Multi-Task Cascaded Convolutional Neural Network (MTCNN) \cite{zhang2016joint} to detect face bounding boxes. Images without detected faces were removed, resulting in a reduction from 162{,}770 to 162{,}127 images in the training set, and from 19{,}866 to 19{,}792 images in the validation set.

For the RVL-CDIP dataset, we used PaddlePaddle Optical Character Recognition Version 3 (PP-OCRv3) \cite{li2022pp} to detect text bounding boxes. Again, images without detected text regions were excluded, reducing the training set from 320{,}000 to 319{,}716 images and the validation set from 40{,}000 to 39{,}972 images. Additionally, we employ four different degrees of anonymization: 25\%, 50\%, 75\%, and 100\%, which refer to the proportion of detected text bounding boxes that are randomly anonymized.

In the case of the internal DOKIQ dataset, all detected text, barcodes, machine-readable zones (MRZ), and faces were anonymized according to the procedure described by Eberhardinger et al. \cite{eberhardinger2025anonymization}, simulating worst-case scenarios in terms of information loss due to anonymization.

Regarding anonymization techniques, we selected three traditional and widely used methods: pixelation, blurring, and masking, as described by Lee et al.~\cite{lee2024balancing}. These methods were chosen due to their relevance in real-world systems and their ease of application across different visual domains. In the context of DOKIQ, no synthetic anonymization module is currently available, making traditional methods the most practical choice. Furthermore, traditional methods enable an easy cross-dataset comparison between facial and document anonymization. Such a comparison would be difficult to conduct with synthetic methods, as it is unclear how to meaningfully compare artificially generated faces with artificially reconstructed text or layout.

Fig.~\ref{fig:anonym_celeba} and~\ref{fig:anonym_rvl_cdip} illustrate example applications of the anonymization techniques on a CelebA person and an RVL-CDIP document, respectively.

\Figure[th](topskip=0pt, botskip=0pt, midskip=0pt)[width=3.3in]{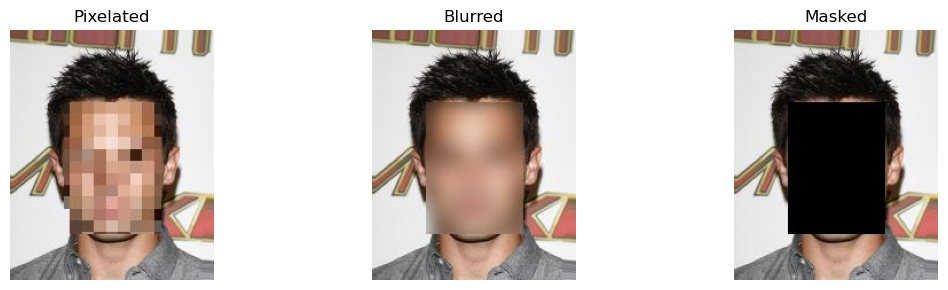}
{\textbf{Anonymized versions of a CelebA person.}\label{fig:anonym_celeba}}
\Figure[th](topskip=0pt, botskip=0pt, midskip=0pt)[width=3.3in]{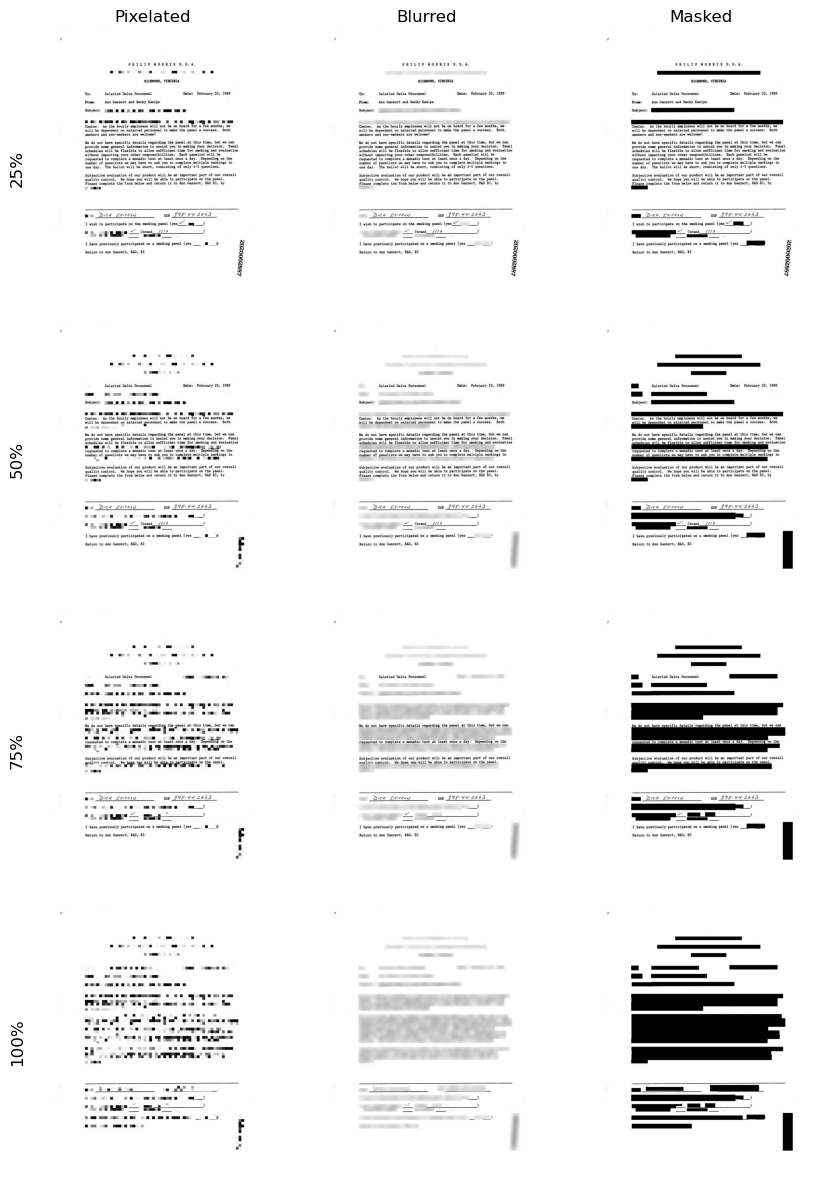}
{\textbf{Anonymized versions of an RVL-CDIP document.}\label{fig:anonym_rvl_cdip}}

\subsection{DINOv2 Training Adaptions}
\label{subsec:din}
In DINOv2's self-distillation paradigm, a student network learns to align the embedding of a crop $\mathbf{u}$ with that of another crop $\mathbf{v}$, seen by a teacher network, using a cross-entropy loss. The authors of Image BERT Pre-training with Online Tokenizer (iBOT) \cite{zhou2021ibot} denote the loss term as follows:
\begin{equation}
    \mathcal{L}_{\text{[CLS]}} = - P^{\text{[CLS]}}_{\theta'}(\mathbf{v})^\top \log P^{\text{[CLS]}}_\theta(\mathbf{u}),
\end{equation}
where $P^{\text{[CLS]}}_{\theta'}(\mathbf{v})$ denotes the teacher's output for crop $\mathbf{v}$, and $P^{\text{[CLS]}}_\theta(\mathbf{u})$ denotes the student's output for crop $\mathbf{u}$. The crops are either local or global. The first are defined by a resolution of 96×96 and cover areas smaller than 50\% of the image. The latter are defined by a resolution of 224×224 and cover areas greater than 50\% of the image. Over these crops, random augmentations are applied, resulting in distorted crops. All crops are passed through the student, while only global crops are passed through the teacher.

We identify, that the two crops $\mathbf{v}$ and $\mathbf{u}$ do not need to originate from the same version of the input image. Instead, $\mathbf{v}$ can be cropped from the original image $x$ and $\mathbf{u}$ from the anonymized version $\hat{x}$ of the same image, or vice versa. This encourages DINOv2 to map the embeddings of original and anonymized versions of same images close to each other in the vector space. Thus, the training dataset $\mathcal{I}$ consisting of original images is augmented by generating anonymized versions $\hat{x}$ for each original image $x \in \mathcal{I}$. These image pairs $(x, \hat{x})$ enable training configurations where crops from both the original and anonymized versions are combined within the DINOv2 training process. In this paper, we explore the following three DINOv2 adaptions:
\begin{itemize}
  \item Adaption A: Global crops are taken from $x$, while local crops are taken from $\hat{x}$.
  \item Adaption B: Global crops are taken from $\hat{x}$, while local crops are taken from $x$.
  \item Adaption C: Both global and local crops are taken from $\hat{x}$.
\end{itemize}

\subsection{Formulation of Retrieval Scenarios}
\label{subsec:for}
Based on sections~\ref{subsec:dat} and~\ref{subsec:din}, we define the set of anonymization methods $\mathcal{A} = \{\text{Pixel}, \text{Blur}, \text{Mask}\}$ and the set of DINOv2 training adaptations $\mathcal{D} = \{\text{A}, \text{B}, \text{C}\}$. For CelebA, these are combined to form $3 \times 3 = 9$ anonymized training configurations. An additional ground-truth model $M_{\text{CelebA-GT}}$ is trained on the original dataset, resulting in the model set:
\[
\{ M_{a,d} \mid a \in \mathcal{A},\ d \in \mathcal{D} \} \cup \{ M_{\text{CelebA-GT}} \}.
\]
For RVL-CDIP, we define the set of anonymization degrees $\mathcal{P} = \{25\%, 50\%, 75\%, 100\%\}$, resulting in $4 \times 3 \times 3 = 36$ anonymized training configurations and the model set:
\[
\{ M_{a,p,d} \mid a \in \mathcal{A},\ p \in \mathcal{P},\ d \in \mathcal{D} \} \cup \{ M_{\text{RVL-CDIP-GT}} \}.
\]
For DOKIQ, we evaluate the worst-case scenarios by applying maximum anonymization, defined as the set $\{\text{max}\}$, across all detected text, barcodes, MRZs, and faces. The resulting model set is defined as:
\[
\{ M_{a,\text{max},d} \mid a \in \mathcal{A},\ d \in \mathcal{D} \} \cup \{ M_{\text{DOKIQ-GT}} \}.
\]

Each model was trained using a ViT-Large backbone\cite{dosovitskiy2020image}. Subsequently, embeddings were generated by inferring the trained models on their respective validation datasets. We denote this as $M(\mathcal{V})$ for inference on the original validation dataset $\mathcal{V}$ and $M(\hat{\mathcal{V}})$ for inference on the anonymized version $\hat{\mathcal{V}}$. For example, $M_{\text{Pixel,A}}(\mathcal{V})$ refers to the embeddings produced by the CelebA model trained with global crops from the original and local crops from the pixelated train dataset on the original validation dataset, while $M_{\text{Pixel,A}}(\hat{\mathcal{V}}_{\text{Pixel}})$ refers to the same model applied to the pixelated validation dataset.

We denote a retrieval scenario as $Q \rightarrow D$, where embeddings from the dataset $Q$ on the left-hand side are used as queries, and embeddings from $D$ on the right-hand side are used as the database. For each image in $Q$, all images from $D$ are retrieved in descending order based on cosine similarity.

Based on these definitions, the original pseudo ground truth retrievals can be defined as follows:
\begin{itemize}
    \item CelebA: $M_{\text{CelebA-GT}}(\mathcal{V}) \rightarrow M_{\text{CelebA-GT}}(\mathcal{V})$
    \item RVL-CDIP: $M_{\text{RVL-CDIP-GT}}(\mathcal{V}) \rightarrow M_{\text{RVL-CDIP-GT}}(\mathcal{V})$
    \item DOKIQ: $M_{\text{DOKIQ-GT}}(\mathcal{V}) \rightarrow M_{\text{DOKIQ-GT}}(\mathcal{V})$
\end{itemize}
These retrieval scenarios represent the performance of the original, non-anonymized systems and are thus used to define the set of $\{\text{relevant documents}\}$ in the context of mAP, as well as the relevance scores $\mathrm{rel}_i$ in the context of mnDCG.

Tab.~\ref{tab:celeba_scenarios},~\ref{tab:rvlcdip_scenarios}, and~\ref{tab:dokiq_scenarios} present the different retrieval scenarios that are evaluated in this paper. Each scenario is assessed against the corresponding pseudo ground truth retrievals defined above. 

\begin{table}[t]
    \centering
    \caption[Retrieval scenarios for CelebA]{Retrieval scenarios for CelebA. }
    \label{tab:celeba_scenarios}
    \begin{tabular}{ll}
        \toprule
        Model & Anonymization method \( a \in \mathcal{A} \) \\
        \midrule
        \multicolumn{2}{c}{\textbf{With original query images}} \\
        \midrule
        Unadapted   & \( M_{\text{CelebA-GT}}(\mathcal{V}) \rightarrow M_{\text{CelebA-GT}}(\hat{\mathcal{V}}_a) \) \\
        Adaption A  & \( M_{a,A}(\mathcal{V}) \rightarrow M_{a,A}(\hat{\mathcal{V}}_a) \) \\
        Adaption B  & \( M_{a,B}(\mathcal{V}) \rightarrow M_{a,B}(\hat{\mathcal{V}}_a) \) \\
        Adaption C  & \( M_{a,C}(\mathcal{V}) \rightarrow M_{a,C}(\hat{\mathcal{V}}_a) \) \\
        \midrule
        \multicolumn{2}{c}{\textbf{With anonymized query images}} \\
        \midrule
        Unadapted   & \( M_{\text{CelebA-GT}}(\hat{\mathcal{V}}_a) \rightarrow M_{\text{CelebA-GT}}(\hat{\mathcal{V}}_a) \) \\
        Adaption A  & \( M_{a,A}(\hat{\mathcal{V}}_a) \rightarrow M_{a,A}(\hat{\mathcal{V}}_a) \) \\
        Adaption B  & \( M_{a,B}(\hat{\mathcal{V}}_a) \rightarrow M_{a,B}(\hat{\mathcal{V}}_a) \) \\
        Adaption C  & \( M_{a,C}(\hat{\mathcal{V}}_a) \rightarrow M_{a,C}(\hat{\mathcal{V}}_a) \) \\
        \bottomrule
    \end{tabular}
\end{table}

\begin{table}[t]
    \centering
    \caption[Retrieval scenarios for RVL-CDIP]{Retrieval scenarios for RVL-CDIP. }
    \label{tab:rvlcdip_scenarios}
    \begin{tabular}{lll}
	\toprule
    Model & \makecell{Anonymization method \( a \in \mathcal{A} \) and \\anonymization degree \( p \in \mathcal{P} \)} \\
    	   \midrule
        \multicolumn{3}{c}{\textbf{With original query images}} \\
        \midrule
        Unadapted   & \( M_{\text{RVL-CDIP-GT}}(\mathcal{V}) \rightarrow M_{\text{RVL-CDIP-GT}}(\hat{\mathcal{V}}_{a,p}) \) & \\
        Adaption A  & \( M_{a,p,A}(\mathcal{V}) \rightarrow M_{a,p,A}(\hat{\mathcal{V}}_{a,p}) \) & \\
        Adaption B  & \( M_{a,p,B}(\mathcal{V}) \rightarrow M_{a,p,B}(\hat{\mathcal{V}}_{a,p}) \) & \\
        Adaption C  & \( M_{a,p,C}(\mathcal{V}) \rightarrow M_{a,p,C}(\hat{\mathcal{V}}_{a,p}) \) & \\
        \midrule
        \multicolumn{3}{c}{\textbf{With anonymized query images}} \\
        \midrule
        Unadapted   & \( M_{\text{RVL-CDIP-GT}}(\hat{\mathcal{V}}_{a,p}) \rightarrow M_{\text{RVL-CDIP-GT}}(\hat{\mathcal{V}}_{a,p}) \) & \\
        Adaption A  & \( M_{a,p,A}(\hat{\mathcal{V}}_{a,p}) \rightarrow M_{a,p,A}(\hat{\mathcal{V}}_{a,p}) \) & \\
        Adaption B  & \( M_{a,p,B}(\hat{\mathcal{V}}_{a,p}) \rightarrow M_{a,p,B}(\hat{\mathcal{V}}_{a,p}) \) & \\
        Adaption C  & \( M_{a,p,C}(\hat{\mathcal{V}}_{a,p}) \rightarrow M_{a,p,C}(\hat{\mathcal{V}}_{a,p}) \) & \\
        \bottomrule
    \end{tabular}
\end{table}

\begin{table}[t]
    \centering
    \caption{Retrieval scenarios for DOKIQ.}
    \label{tab:dokiq_scenarios}
    \begin{tabular}{ll}
        \toprule
        Model & Anonymization method \( a \in \mathcal{A} \) \\
        \midrule
        \multicolumn{2}{c}{\textbf{With original query images}} \\
        \midrule
        Unadapted   & \( M_{\text{DOKIQ-GT}}(\mathcal{V}) \rightarrow M_{\text{DOKIQ-GT}}(\hat{\mathcal{V}}_{a,\text{max}}) \) \\
        Adaption A  & \( M_{a,\text{max},A}(\mathcal{V}) \rightarrow M_{a,\text{max},A}(\hat{\mathcal{V}}_{a,\text{max}}) \) \\
        Adaption B  & \( M_{a,\text{max},B}(\mathcal{V}) \rightarrow M_{a,\text{max},B}(\hat{\mathcal{V}}_{a,\text{max}}) \) \\
        Adaption C  & \( M_{a,\text{max},C}(\mathcal{V}) \rightarrow M_{a,\text{max},C}(\hat{\mathcal{V}}_{a,\text{max}}) \) \\
        \midrule
        \multicolumn{2}{c}{\textbf{With anonymized query images}} \\
        \midrule
        Unadapted   & \( M_{\text{DOKIQ-GT}}(\hat{\mathcal{V}}_{a,\text{max}}) \rightarrow M_{\text{DOKIQ-GT}}(\hat{\mathcal{V}}_{a,\text{max}}) \) \\
        Adaption A  & \( M_{a,\text{max},A}(\hat{\mathcal{V}}_{a,\text{max}}) \rightarrow M_{a,\text{max},A}(\hat{\mathcal{V}}_{a,\text{max}}) \) \\
        Adaption B  & \( M_{a,\text{max},B}(\hat{\mathcal{V}}_{a,\text{max}}) \rightarrow M_{a,\text{max},B}(\hat{\mathcal{V}}_{a,\text{max}}) \) \\
        Adaption C  & \( M_{a,\text{max},C}(\hat{\mathcal{V}}_{a,\text{max}}) \rightarrow M_{a,\text{max},C}(\hat{\mathcal{V}}_{a,\text{max}}) \) \\
        \bottomrule
    \end{tabular}
\end{table}
The motivation behind the ``\textbf{With anonymized query images}'' scenarios is to investigate whether anonymizing the query image can help mitigate the domain shift introduced by anonymization. By matching the visual characteristics of the query image to those of the anonymized database images, semantic similarity may be restored, potentially improving retrieval quality despite the loss of information through anonymization. In addition, it simulates the DOKIQ use case of comparing closed cases as described in section~\ref{subsec:dok}.

\subsection{Downstream Tasks}
Our hypothesis is that models which produce retrieval results most similar to the original retrievals also learn the highest-quality embeddings for downstream classification tasks. 
The rationale behind this hypothesis is that retrieval results directly reflect the structure of the learned embedding space. If two models return very similar retrieval results for the same query images, this suggests that they capture comparable semantic relationships. Since downstream classification tasks rely on the same separability of vectors, which belong to distinct classes, we expect that those models, which retain the distance-relations of the ground-truth model will also perform better in classification.

To test this, we perform downstream evaluations on the anonymized public datasets: gender classification on CelebA and document classification on RVL-CDIP. Subsequently, we compute Pearson Correlation Coefficient (PCC)s between the retrieval metrics and the classification accuracies to quantify the relationship between retrieval quality and downstream task performance.

\subsection{Attention Maps}
To enable a qualitative analysis of model behavior, we visualize attention maps \cite{guo2022attention} produced by DINOv2. Attention maps highlight image regions deemed semantically important by the model during training and inference. If such high-attention areas are obscured by anonymization, the resulting information loss may help explain performance drops in retrieval or downstream classification.

\section{Results}
\label{sec:res}
\subsection{Quantitative Results}
\subsubsection{CelebA}
Tab.~\ref{tab:celeba_retrieval_qualities} shows the retrieval qualities of all scenarios on CelebA. As one can see, the
best scenario is Unadapted/Pixel with original query images and the worst retrieval qualities were produced by Adaption A/Mask and Adaption B/Mask.
\begin{table}[t]
    \centering
    \caption{CelebA retrieval qualities.}
    \label{tab:celeba_retrieval_qualities}
    \begin{tabular}{@{}llcccclccc@{}}
        \toprule
        & & \multicolumn{3}{c}{mAP} & & \multicolumn{3}{c}{mnDCG\textsubscript{989}} \\
        \cmidrule(lr){3-5} \cmidrule(lr){7-9}
        Model & & Pixel & Blur & Mask & & Pixel & Blur & Mask \\
        \midrule
        \multicolumn{9}{c}{\textbf{With original query images}} \\
        \midrule
        Unadapted & & \textbf{73.2} & 67.3  & 31.0 & & \textbf{94.2} & 92.6 & 77.1 \\
        Adaption A & & 57.0 & 50.6 & 13.7 & & 89.3 & 86.8 & 64.2 \\
        Adaption B & & 38.5 &50.8 & 13.6 & & 81.6 & 86.9 & 64.1 \\
        Adaption C & & 45.4 & 23.8 & 10.2 & & 84.7 & 71.3 & 58.6 \\
        \midrule
        \multicolumn{9}{c}{\textbf{With anonymized query images}} \\
        \midrule
        Unadapted & & \textbf{67.6} & 62.4 & 35.8 & & \textbf{92.7} & 91.1 & 79.9 \\
        Adaption A & & 56.3 & 55.0 & 17.4 & & 89.0 & 88.4 & 67.7 \\
        Adaption B & & 38.5 & 53.7 & 16.4 & & 81.4 & 88.0 & 67.3 \\
        Adaption C & & 49.1 & 52.5 & 27.4 & & 86.2 & 87.6 & 75.1 \\
        \bottomrule
    \end{tabular}
\end{table}
On the downstream task listed in Tab.~\ref{tab:celeba_gender_detection_accuracy}, the best performing model was Adaption C/Blur, i.e., the DINOv2 model trained solely on blurred faces. The worst performing models were Adaption A/Mask and Adaption B/Mask, reflecting their bad retrieval performance.
\begin{table}[t]
    \centering
    \caption{CelebA gender detection accuracies.}
    \label{tab:celeba_gender_detection_accuracy}
    \begin{tabular}{@{}llcccclccc@{}}
        \toprule
        & & \multicolumn{3}{c}{k-NN} & & \multicolumn{3}{c}{Linear} \\
        \cmidrule(lr){3-5} \cmidrule(lr){7-9}
        Model & & Pixel & Blur & Mask & & Pixel & Blur & Mask \\
        \midrule
        Unadapted & & 87.9 & 86.3 & 84.2 & & 90.6 & 89.9 & 85.5 \\
        Adaption A & & 85.9 & 86.3 & 71.1 & & 89.4 & 89.7 & 77.3 \\
        Adaption B & & 80.4 & 85.6 & 71.6 & & 84.4 & 88.7 & 77.9 \\
        Adaption C & & 85.1 & \textbf{89.8} & 78.9 & & 89.0 & \textbf{93.4} & 82.8 \\
        \bottomrule
    \end{tabular}
\end{table}
Fig.~\ref{fig:celeba_corr} shows the correlation between all CelebA metrics. The highest PCCs between retrieval metrics and downstream task performance can be found in the row mnDCG (anonymized), with PCCs of 0.95. This suggests that performing the mnDCG evaluation with anonymized query images is the most reliable predictor of downstream task performance among the metrics examined.

\Figure[th](topskip=0pt, botskip=0pt, midskip=0pt)[width=3.3in]{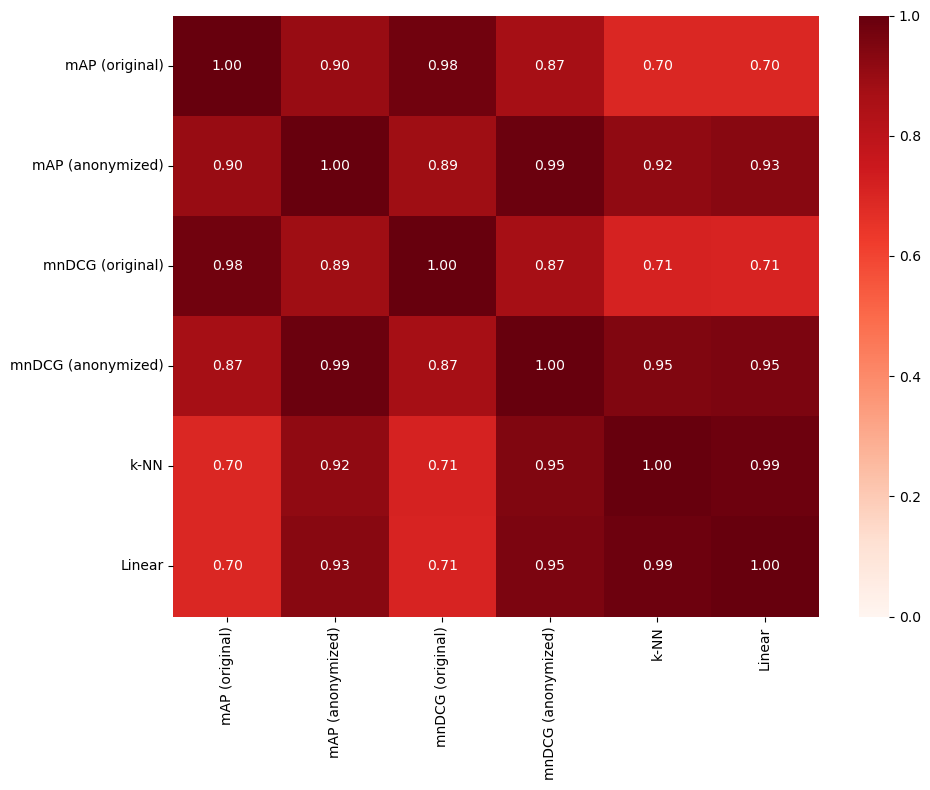}
{\textbf{CelebA metrics correlation matrix. "(original)" and "(anonymized)" refer to the query images used.}\label{fig:celeba_corr}}

\subsubsection{RVL-CDIP}
For RVL-CDIP, we present results for 25\% and 100\% anonymization to illustrate the two extremes of minimum and maximum information loss. Results for the anonymization degrees 50\% and 75\% are provided in the appendix.

Following the detailed results, we also present summary plots that visualize the performance trends across different anonymization degrees in a compact and interpretable manner.

Tab.~\ref{tab:rvl_cdip_retrieval_qualities_at_25} shows the retrieval qualities of all scenarios on RVL-CDIP with an anonymization degree of 25\%. Again, the Unadapted model performed the best. However, this time in combination with the blurring method. The worst performing scenarios were Adaption A/Mask and Adaption C/Mask.
\begin{table}[t]
    \centering
    \caption{RVL-CDIP retrieval qualities @ 25\% anonymization.}
    \label{tab:rvl_cdip_retrieval_qualities_at_25}
    \begin{tabular}{@{}llcccclccc@{}}
        \toprule
        & & \multicolumn{3}{c}{mAP} & & \multicolumn{3}{c}{mnDCG\textsubscript{1998}} \\
        \cmidrule(lr){3-5} \cmidrule(lr){7-9}
        Model & & Pixel & Blur & Mask & & Pixel & Blur & Mask \\
        \midrule
        \multicolumn{9}{c}{\textbf{With original query images}} \\
        \midrule
        Unadapted & & 71.5 & \textbf{85.4} & 55.4 & & 93.2 & \textbf{97.2} & 87.5 \\
        Adaption A & & 57.2 & 60.7 & 23.2 & & 88.8 & 90.1 & 68.6 \\
        Adaption B & & 53.3 & 59.0 & 38.0 & & 87.3 & 89.5 & 79.3 \\
        Adaption C & & 53.9 & 59.6 & 28.3 & & 87.5 & 89.7 & 72.9 \\
        \midrule
        \multicolumn{9}{c}{\textbf{With anonymized query images}} \\
        \midrule
        Unadapted & & 66.5 & \textbf{78.6} & 55.1 & & 91.7 & \textbf{95.5} & 87.5 \\
        Adaption A & & 56.1 & 59.6 & 28.0 & & 88.3 & 89.7 & 72.7 \\
        Adaption B & & 52.2 & 58.8 & 44.0 & & 86.8 & 89.4 & 83.0 \\
        Adaption C & & 54.5 & 59.4 & 36.8 & & 87.7 & 89.6 & 79.0 \\
        \bottomrule
    \end{tabular}
\end{table}
Tab.~\ref{tab:rvl_cdip_document_classification_accuracy_at_25} shows the downstream task performances. This time, Adaption A/Blur achieved the highest accuracy, i.e., the DINOv2 model trained with global crops from original and local crops from blurred documents. Adaption A/Mask and Adaption C/Mask, which performed the worst in image retrieval, also achieved the lowest accuracies.
\begin{table}[t]
    \centering
    \caption{RVL-CDIP document classification accuracies @ 25\% anonymization.}
    \label{tab:rvl_cdip_document_classification_accuracy_at_25}
    \begin{tabular}{@{}llcccclccc@{}}
        \toprule
        & & \multicolumn{3}{c}{k-NN} & & \multicolumn{3}{c}{Linear} \\
        \cmidrule(lr){3-5} \cmidrule(lr){7-9}
        Model & & Pixel & Blur & Mask & & Pixel & Blur & Mask \\
        \midrule
        Unadapted & & 64.7 & 67.8 & 62.7 & & 60.2 & 62.4 & 59.9 \\
        Adaption A & & 67.1 & \textbf{68.0} & 47.0 & & 62.0 & \textbf{62.8} & 50.5 \\
        Adaption B & & 63.9 & 67.6 & 62.7 & & 59.8 & 62.7 & 60.0 \\
        Adaption C & & 65.5 & 67.6 & 57.7 & & 60.9 & 62.0 & 57.9 \\
        \bottomrule
    \end{tabular}
\end{table}
Fig.~\ref{fig:rvl_cdip_25_corr_matrix} shows that mnDCG with anonymized query images has again the highest PCCs with downstream task performances.

\Figure[th](topskip=0pt, botskip=0pt, midskip=0pt)[width=3.3in]{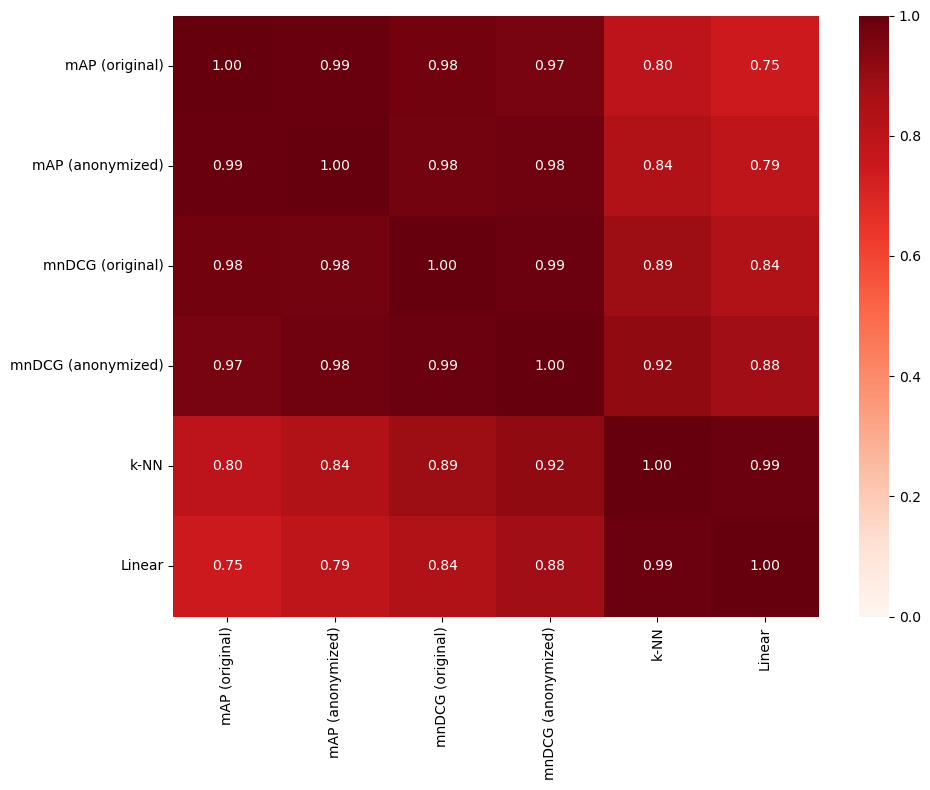}
{\textbf{RVL-CDIP metrics correlation matrix @ 25\% anonymization.}\label{fig:rvl_cdip_25_corr_matrix}}
For the 100\% anonymization degree, the best retrieval performances were achieved again with Unadapted/Blur. In contrast to the other anonymization degrees, this time both mAP and mnDCG slightly favoured anonymizing the query images, as shown in Tab.~\ref{tab:rvl_cdip_retrieval_qualities_at_100}.
\begin{table}[t]
    \centering
    \caption{RVL-CDIP retrieval qualities @ 100\% anonymization.}
    \label{tab:rvl_cdip_retrieval_qualities_at_100}
    \begin{tabular}{@{}llcccclccc@{}}
        \toprule
        & & \multicolumn{3}{c}{mAP} & & \multicolumn{3}{c}{mnDCG\textsubscript{1998}} \\
        \cmidrule(lr){3-5} \cmidrule(lr){7-9}
        Model & & Pixel & Blur & Mask & & Pixel & Blur & Mask \\
        \midrule
        \multicolumn{9}{c}{\textbf{With original query images}} \\
        \midrule
        Unadapted & & 34.6 & \textbf{50.5} & 28.0 & & 75.4 & \textbf{85.7} & 71.7 \\
        Adaption A & & 16.3 & 39.9 & 6.7 & & 62.4 & 80.2 & 51.4 \\
        Adaption B & & 24.2 & 17.8 & 18.0 & & 67.4 & 62.3 & 61.2 \\
        Adaption C & & 22.5 & 16.2 & 14.8 & & 66.2 & 58.7 & 57.4 \\
        \midrule
        \multicolumn{9}{c}{\textbf{With anonymized query images}} \\
        \midrule
        Unadapted & & 38.5 & \textbf{50.6} & 35.7 & & 79.3 & \textbf{85.8} & 78.2 \\
        Adaption A & & 21.1 & 49.3 & 10.6 & & 67.7 & 85.4 & 58.5 \\
        Adaption B & & 32.1 & 27.4 & 31.0 & & 76.1 & 74.0 & 75.6 \\
        Adaption C & & 33.3 & 25.0 & 33.5 & & 77.0 & 71.4 & 77.3 \\
        \bottomrule
    \end{tabular}
\end{table}
In Tab.~\ref{tab:rvl_cdip_document_classification_accuracy_at_100}, one can see that Unadapted/Blur was the best performing model on the downstream task, in contrast to all other anonymization degrees, where Adaption A/Blur achieved the highest accuracies.
\begin{table}[t]
    \centering
    \caption{RVL-CDIP document classification accuracies @ 100\% anonymization.}
    \label{tab:rvl_cdip_document_classification_accuracy_at_100}
    \begin{tabular}{@{}llcccclccc@{}}
        \toprule
        & & \multicolumn{3}{c}{k-NN} & & \multicolumn{3}{c}{Linear} \\
        \cmidrule(lr){3-5} \cmidrule(lr){7-9}
        Model & & Pixel & Blur & Mask & & Pixel & Blur & Mask \\
        \midrule
        Unadapted & & 55.2 & \textbf{67.8} & 58.3 & & 54.1 & \textbf{62.6} & 56.0 \\
        Adaption A & & 40.1 & 66.0 & 34.9 & & 45.1 & 61.8 & 40.8 \\
        Adaption B & & 52.1 & 56.8 & 59.8 & & 52.9 & 55.7 & 56.3 \\
        Adaption C & & 54.4 & 52.5 & 61.9 & & 55.0 & 54.0 & 57.6 \\
        \bottomrule
    \end{tabular}
\end{table}
Fig.~\ref{fig:rvl_cdip_100_corr_matrix} again shows that mnDCG computed with anonymized query images exhibits the highest PCCs with downstream performances. However, in contrast to previous correlation matrices, we observe that for the ``\textbf{With original query images}'' scenarios, mAP yielded higher PCCs than mnDCG.

\Figure[th](topskip=0pt, botskip=0pt, midskip=0pt)[width=3.3in]{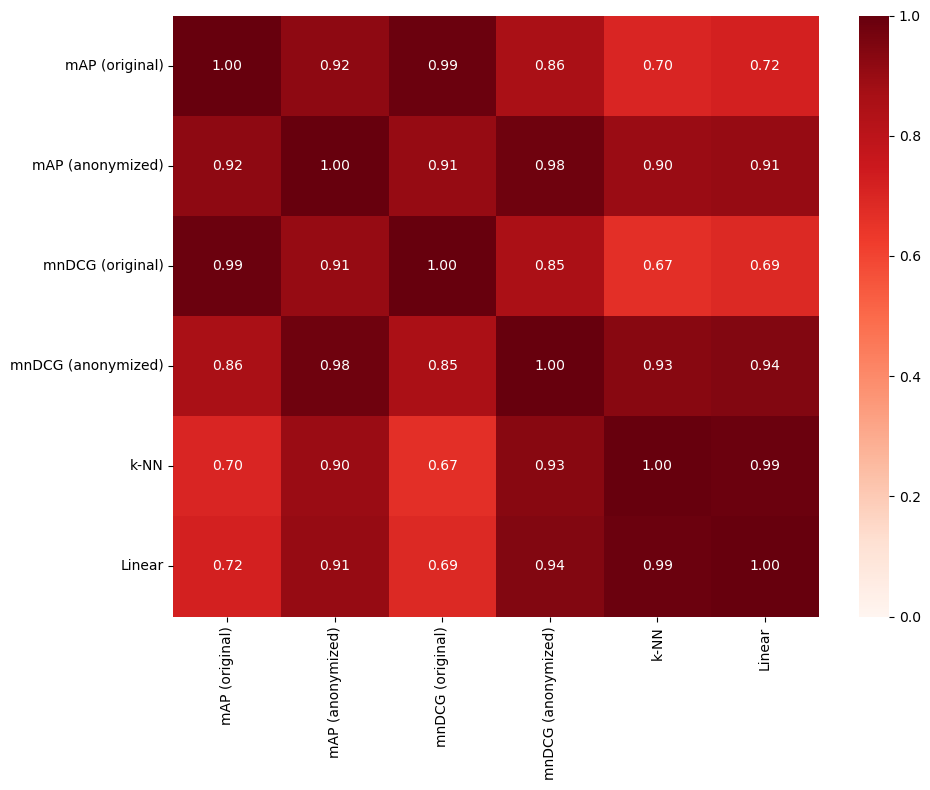}
{\textbf{RVL-CDIP metrics correlation matrix @ 100\% anonymization.}\label{fig:rvl_cdip_100_corr_matrix}}

\Figure[th](topskip=0pt, botskip=0pt, midskip=0pt)[width=3.3in]{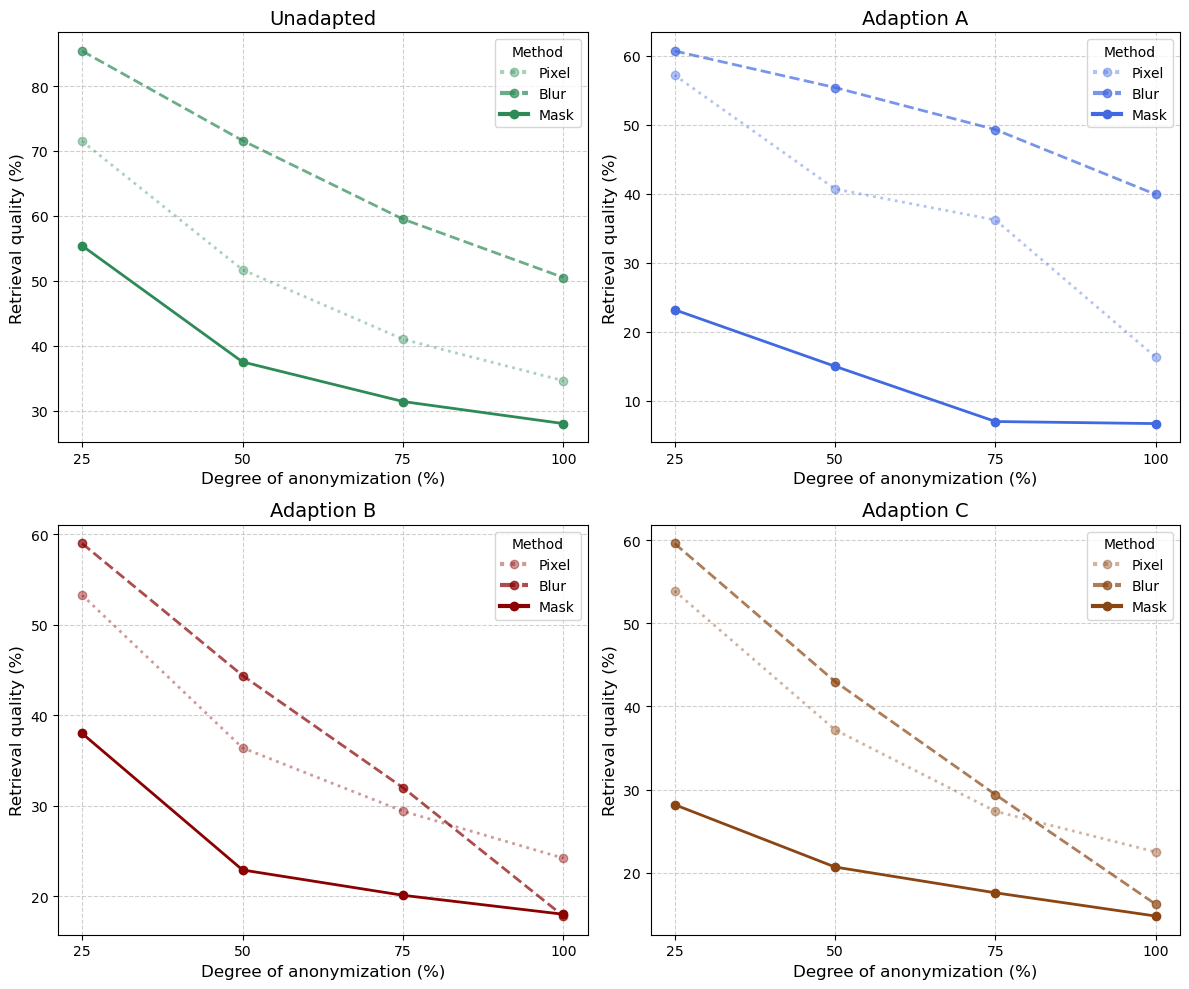}
{\textbf{mAP of the different RVL-CDIP-DINOv2 models with original query images and different degrees of anonymization. The figure shows a clear negative trend in retrieval quality across all model types and anonymization methods as the degree of anonymization increases.}\label{fig:impoq}}

\Figure[th](topskip=0pt, botskip=0pt, midskip=0pt)[width=3.3in]{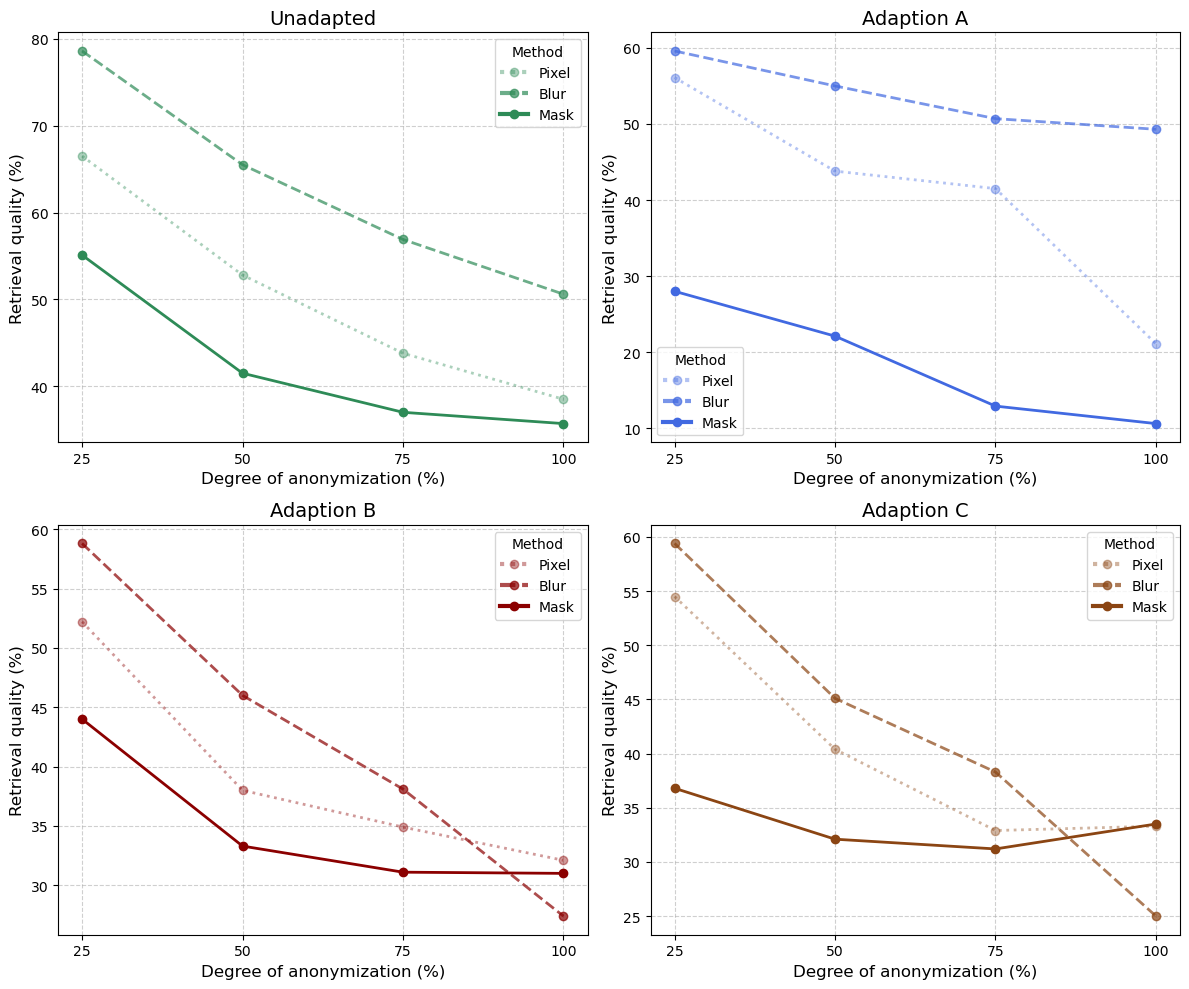}
{\textbf{mAP of the different RVL-CDIP-DINOv2 models with anonymized query images and different degrees of anonymization. The figure shows an overall negative trend in retrieval quality for most model types and anonymization methods as the degree of anonymization increases, although some training configurations — such as Adaption C/Mask — exhibit the opposite behavior.}}\label{fig:impaq}

\Figure[th](topskip=0pt, botskip=0pt, midskip=0pt)[width=3.3in]{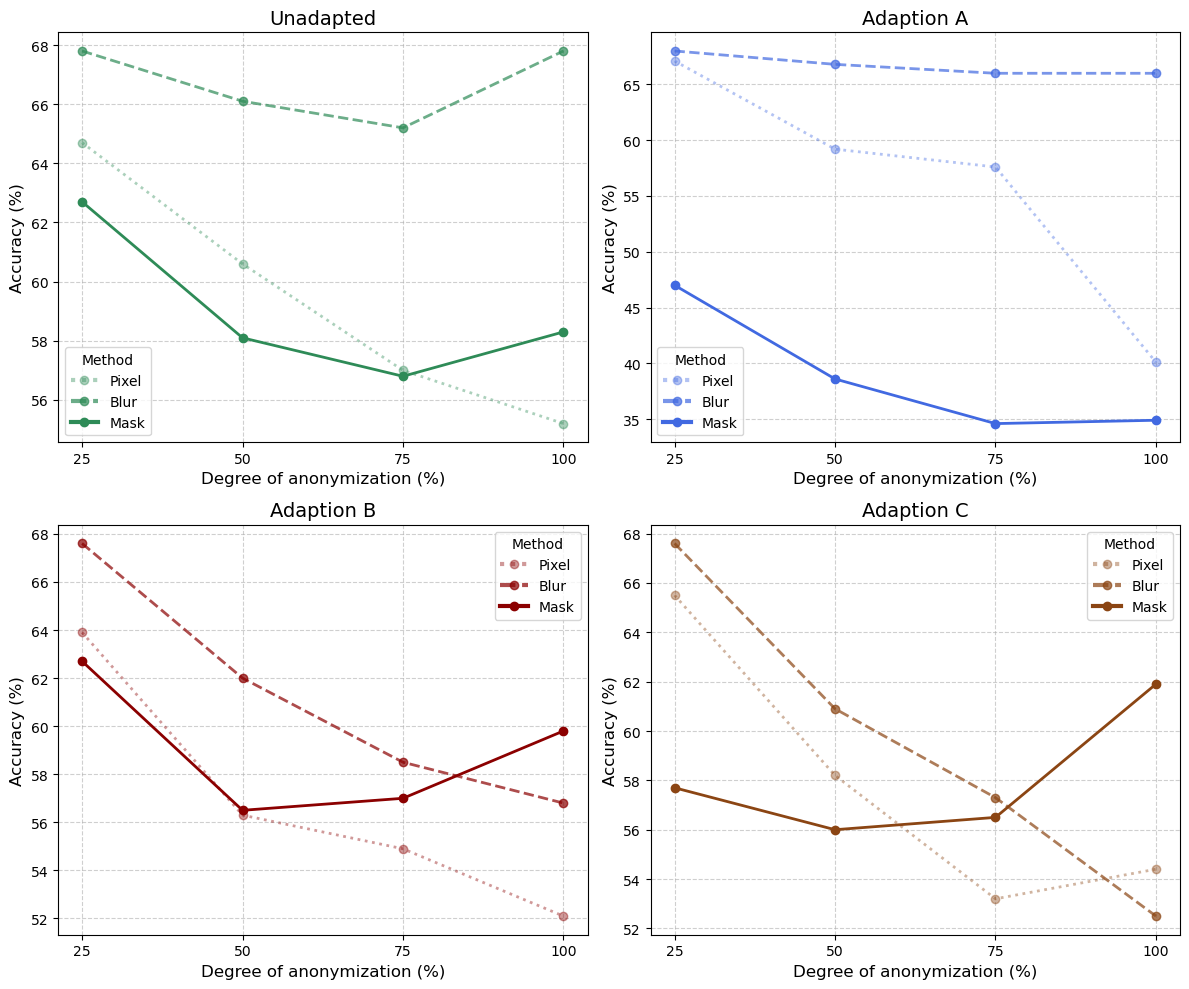}
{\textbf{k-NN accuracies of the different RVL-CDIP-DINOv2 models for different degrees of anonymization. This figure shows that, in contrast to the retrieval results, several training configurations exhibit a positive trend between the degree of anonymization and classification accuracy.}\label{fig:impknn}}

Fig.~\ref{fig:impoq} shows the impact of different degrees of anonymization on the mAP performance when querying with original images. As one can see, the retrieval quality degrades across all models and anonymization methods with increasing degrees of anonymization. We observe the same for the mnDCG retrieval qualities.

Fig.~\ref{fig:impaq} shows the impact of different degrees of anonymization on the mAP performance when querying with anonymized images. Interestingly, this time there are cases in which the retrieval quality increases with an
increasing degree of anonymization or does not decrease any further. This is the case, for example, with Adaption C/Mask, where a 100\% degree of anonymization led to better performance than with 50\% or 75\% anonymization. We observe a similar pattern for the respective mnDCG retrieval qualities.

Fig.~\ref{fig:impknn} shows the impact of different degrees of anonymization on the k-NN performance. The trend that emerged in Fig.~\ref{fig:impaq} is even more pronounced this time. In 6 out of 12 cases, performance has improved with 100\% anonymization. It is noteworthy that 3 of these 6 cases are the mask anonymization method. We observe a similar pattern for linear classification performance.

\subsubsection{DOKIQ}
Tab.~\ref{tab:dokiq_retrieval_qualities} shows the retrieval qualities of all scenarios on DOKIQ. In the ``\textbf{With original query images}'' scenarios, the best performance was achieved by Unadapted/Pixel. Interestingly, in the ``\textbf{With anonymized query images}'' scenarios, Adaption B/Pixel yielded the highest performance, at least when evaluated using mnDCG. This marks the only case across all evaluated retrieval scenarios where the Unadapted model was not the top-performing system.
\begin{table}[t]
    \centering
    \caption[DOKIQ retrieval qualities]{DOKIQ retrieval qualities.}
    \label{tab:dokiq_retrieval_qualities}
    \begin{tabular}{@{}llcccclccc@{}}
        \toprule
        & & \multicolumn{3}{c}{mAP} & & \multicolumn{3}{c}{mnDCG\textsubscript{750}} \\
        \cmidrule(lr){3-5} \cmidrule(lr){7-9}
        Model & & Pixel & Blur & Mask & & Pixel & Blur & Mask \\
        \midrule
        \multicolumn{9}{c}{\textbf{With original query images}} \\
        \midrule
        Unadapted & & \textbf{65.7} & 50.8 & 43.2 & & \textbf{92.9} & 88.2 & 85.4 \\
        Adaption A & & 45.6 & 43.7 & 36.9 & & 87.9 & 86.8 & 83.9 \\
        Adaption B & & 46.1 & 43.9 & 37.5 & & 88.0 & 87.1 & 84.1 \\
        Adaption C & & 42.8 & 36.8 & 33.9 & & 86.4 & 83.1 & 81.6 \\
        \midrule
        \multicolumn{9}{c}{\textbf{With anonymized query images}} \\
        \midrule
        Unadapted & & \textbf{48.0} & 40.9 & 34.2 & & 87.7 & 85.2 & 82.4 \\
        Adaption A & & 44.4 & 43.5 & 36.5 & & 87.5 & 87.1 & 84.0 \\
        Adaption B & & 46.5 & 45.6 & 39.9 & & \textbf{88.4} & 88.0 & 85.6 \\
        Adaption C & & 46.3 & 44.1 & 40.8 & & 88.2 & 87.4 & 86.0 \\
        \bottomrule
    \end{tabular}
\end{table}

\subsection{Qualitative Results: Attention Maps}
To qualitatively assess which regions of the input images are most relevant to the model, we visualize attention maps generated by the two models $M_{\text{CelebA-GT}}$ and $M_{\text{RVL-CDIP-GT}}$ on the original validation sets. In the visualizations, brighter regions indicate higher attention values. The attention maps were extracted from the last layer of all 16 heads of the ViT-Large backbones trained during the study.

\Figure[th](topskip=0pt, botskip=0pt, midskip=0pt)[width=3.3in]{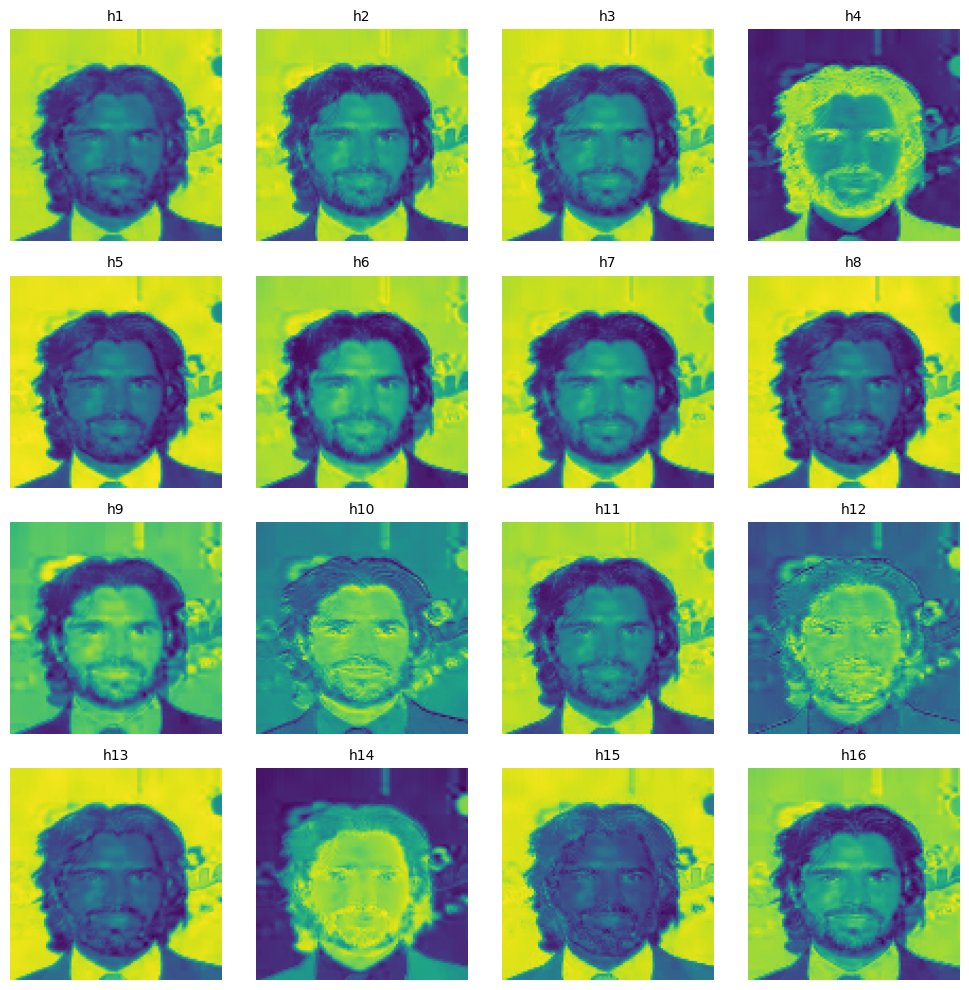}
{\textbf{Attention map visualization of a male CelebA person. It is visible that different heads attend to distinct and semantically meaningful regions of the image.}\label{fig:att_male}}
Fig.~\ref{fig:att_male} shows the attention maps of a male CelebA person of the validation dataset, as produced by $M_{\text{CelebA-GT}}$. It becomes evident that the semantic information within the image is distributed across different regions, with various heads focusing on distinct aspects. For example, the background appears to be particularly important for certain heads, such as h13 and h15. Other heads, like h4, focus more strongly on hair features, while h14 shows increased attention in facial regions. Naturally, masking the face removes all semantic content from this region, which can heavily impact performance when several heads rely on facial features.

\Figure[th](topskip=0pt, botskip=0pt, midskip=0pt)[width=3.3in]{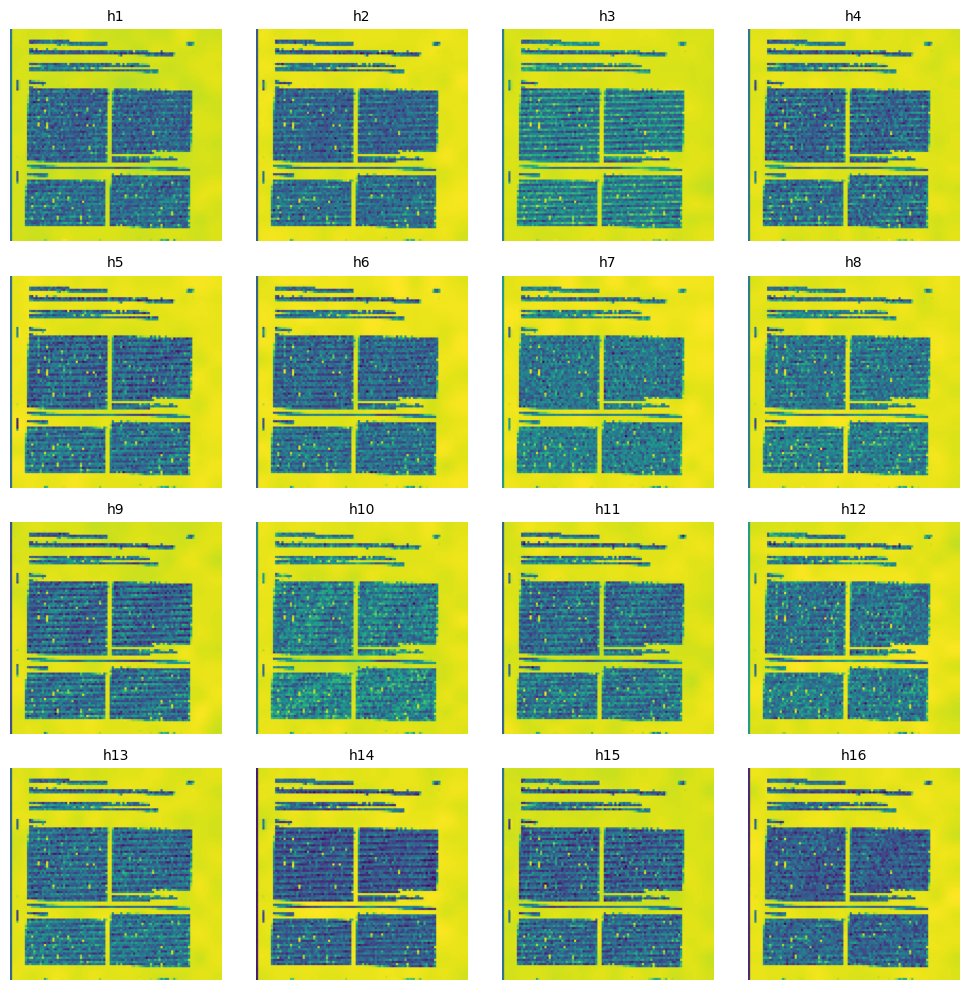}
{\textbf{Attention map visualization of an RVL-CDIP document of the category scientific publication. In contrast to images of faces, all attention heads primarily attend to background regions.}\label{fig:att_sci}}
Fig.~\ref{fig:att_sci} shows the attention maps of an RVL-CDIP scientific publication document of the validation dataset, as produced by $M_{\text{RVL-CDIP-GT}}$. Across all 16 heads, the model predominantly attends to the background areas of the document, i.e., the white space. In contrast, the textual regions receive noticeably lower attention, as indicated by the darker areas in the maps.

This observation suggests that the semantic understanding of DINOv2 is largely based on the spatial layout and distribution of non-textual regions, rather than on the text content itself, unlike dedicated Optical Character Recognition (OCR)-based models. As a result, masking the text does not necessarily disrupt the structural semantics of a document. In fact, at high anonymization degrees (e.g., 100\%), masking may even be advantageous, as it preserves the underlying layout, cf. Fig.~\ref{fig:impaq} and Fig. \ref{fig:impknn}. This may explain why masking, in some cases, outperforms pixelation, cf. Tab.~\ref{tab:rvl_cdip_document_classification_accuracy_at_100}.

\section{Discussion}
\label{sec:dis}

\subsection{Impact of Different Anonymization Methods}
Across all experiments, masking never achieved the best performance, neither in retrieval nor in downstream classification. In contrast, the performance of pixelation and blurring varied depending on the dataset: pixelation performed best on the DOKIQ dataset, while blurring yielded superior results on RVL-CDIP. On the CelebA dataset, it is more nuanced: while pixelation led to the best retrieval performance, blurring resulted in the highest accuracy for gender classification. This suggests that the optimal anonymization method is task-specific, even within the same dataset.

Nevertheless, masking should not be disregarded entirely. In scenarios where global structures remain visually intact despite masking, such as the layout of documents, this method can still preserve semantic cues. This may be particularly beneficial at high anonymization degrees (e.g., 100\%), where other methods may introduce visual noise.

\subsection{Impact of Different Anonymization Degrees}
On RVL-CDIP, we observe a clear negative trend between the anonymization degree and retrieval quality: the more image regions are anonymized, the lower the retrieval performance. This supports research efforts aiming to minimize the extent of anonymization while still ensuring privacy-compliance ~\cite{eberhardinger2025anonymization}. In the case of DOKIQ, these results suggest that minimizing the anonymization degree, where legally permissible, should significantly enhance retrieval quality.

However, for downstream tasks such as document classification, this relationship is not strictly monotonic. On RVL-CDIP, we find that fully anonymizing all detected text regions can in some cases outperform partial anonymization. One possible explanation is that partial anonymization introduces visual inconsistencies, while full anonymization restores structural coherence.

\subsection{Training on Anonymized Data}
A key finding of our work is that the effectiveness of training on anonymized data depends on the specific task. For all retrieval evaluations on the public datasets, the Unadapted models consistently achieved the best performance. An exception is DOKIQ, where Adaption B led to the highest mnDCG scores when querying with anonymized images. A possible explanation for this behaviour is discussed in section~\ref{subsec:bia}.

In contrast, the downstream classification results were more heterogeneous. In some cases, the Unadapted model performed best. In others, one of the adapted models outperformed it.

The fact that Adaption A and B led to the highest performance in some cases also serves as a proof of concept for the proposed DINOv2 training adaptions outlined in section~\ref{subsec:din}.

\subsection{Relationship Between Retrieval and Downstream Task Performance}
As shown in all correlation matrices, the metric that most strongly correlates with downstream task performance is mnDCG when using anonymized query images. This finding suggests that mnDCG might be better suited than mAP to capture the underlying embedding quality. Unlike mAP, which is based on a binary notion of relevance, mnDCG incorporates the cosine similarity values from the original system as relevance scores. As a result, it reflects not only whether relevant documents are retrieved, but also how closely the embedding space of the evaluated systems align with that of the ground truth system. This makes mnDCG particularly valuable for evaluating model performance under domain shift caused by anonymization.

\subsection{Retrieval Bias}
The consistently strong performance of the Unadapted models in the retrieval evaluations may be attributed to an inherent retrieval bias. Since the ground truth retrievals were defined using the same models, $M_{\text{CelebA-GT}}$, $M_{\text{RVL-CDIP-GT}}$, and $M_{\text{DOKIQ-GT}}$, that are also used during evaluation, their produced embeddings are naturally more aligned with themselves. That is, the same network with identical weights processes nearly identical pixel inputs, only differing in the anonymized regions, potentially introducing a systematic advantage.

Nevertheless, this finding remains meaningful: it demonstrates that the original models, trained on non-anonymized data, tend to produce the most similar retrievals after anonymization. This suggests that retrieval behavior is largely preserved despite anonymizing the input, unless the anonymization degree is particularly severe, as seen in the DOKIQ worst case scenarios. In this case, at least for the ``\textbf{With anonymized query images}'' scenarios, the retrieval bias was substantially reduced because the images differ significantly from those of the original validation dataset due to maximum anonymization. This pattern is reflected in the 100\% anonymization setting of the RVL-CDIP dataset: in the ``\textbf{With anonymized query images}'' scenarios, Adaption A achieved a retrieval performance close to the Unadapted model, cf. Tab.~\ref{tab:rvl_cdip_retrieval_qualities_at_100}. In contrast, with only 25\% anonymization, the performance gap between adapted and unadapted models was noticeably larger, cf. Tab.~\ref{tab:rvl_cdip_retrieval_qualities_at_25}. This highlights how the degree of anonymization impacts retrieval bias.

\label{subsec:bia}
\subsection{Practical Recommendations}
\subsubsection{Keep the Original Embeddings}
A seemingly simple approach would be to use the original embeddings and only anonymize the images. The retrieval results would be identical to those before anonymization as long as the association between anonymized images and the corresponding original embeddings is intact. However, the legal admissibility of this approach is unclear. It remains uncertain whether storing embeddings produced on non-anonymized data complies with GDPR. This uncertainty is further amplified by research on the reversibility of embeddings~\cite{mahendran2015understanding}, which raises additional concerns regarding potential restoring of PII.

\subsubsection{Keep the Original Model}
\label{sec:keep-model}
If retaining the original embeddings is not legally permissible, keeping the original model is very likely beneficial. A strong bias was observed in favor of the Unadapted models during retrieval evaluations. Reusing them to infer on anonymized data yielded the most consistent retrieval results when compared to newly trained models. However, to fully benefit from this retrieval bias, the anonymized image regions should be kept as small as possible, while still complying with data protection regulations.

\subsubsection{Evaluate on a Case-By-Case Basis}
The methods introduced in this paper can serve as a general evaluation framework to evaluate the impact of anonymization on image retrieval on a case-by-case basis. Given a model $M_{\text{GT}}$, trained on the original data, and a validation dataset $\mathcal{V}$ that is yet to be anonymized, the best possible retrievals can be defined as:

\[
M_{\text{GT}}(\mathcal{V}) \rightarrow M_{\text{GT}}(\mathcal{V}),
\] using the notation introduced in section~\ref{subsec:for}. These ground truth retrievals form the basis for evaluation: they define the set of $\{ \text{relevant documents} \}$ for computing mAP, and provide the relevance scores $\mathrm{rel}_i$ for mnDCG via cosine similarities. All other retrieval scenarios, regardless of the backbone architecture, training adaptions, or anonymization techniques, can then be evaluated against this ground truth. This makes the framework flexible and extensible, allowing researchers and engineers to quantify how closely their anonymized systems approximate the performance of the original system.

\subsection{Ethical Considerations}
From an ethical and societal perspective, our results show a trade-off between retrieval performance and data protection. We found that retrieval quality benefits from keeping anonymization to a minimum. As explained in section \ref{sec:keep-model} anonymized areas should be kept as small as possible in order to take full advantage of retrieval bias while complying with regulations. Similar trade-offs have been described in the literature on dense prediction tasks. 
However, this finding could inadvertently encourage organizations to reduce anonymization excessively, creating the risk of non-compliance with privacy regulations. Therefore, our results underscore the need to carefully balance model performance against regulatory compliance and ethical considerations, and to focus future research on methods that maintain strong computer vision performance even under strict anonymization conditions.

\subsection{Outlook}
Future work could explore synthetic anonymization techniques in the context of image retrieval. In DOKIQ, for example, this could involve the generation of artificial text, barcodes, MRZs, and faces in order to preserve the visual semantics of documents while removing PII. Of course, the specific design of such approaches depends on the dataset at hand and the type of entities to be anonymized.

Additionally, improvements to traditional methods may further preserve model performance. For example, Shvai et al.~\cite{shvai2023adaptive} introduced a learnable network that modifies blurred areas such that the original network still classifies blurred images correctly. The evaluation framework proposed in this paper provides a basis to systematically compare such advanced anonymization techniques in terms of their retrieval utility.

\appendix
\section{\break RVL-CDIP Performance Tables and Correlation Matrices}
\begin{table}[ht]
    \centering
    \caption[RVL-CDIP retrieval qualities @ 50\% anonymization]{RVL-CDIP retrieval qualities @ 50\% anonymization.}
    \label{tab:rvl_cdip_retrieval_qualities_at_50}
    \begin{tabular}{@{}llcccclccc@{}}
        \toprule
        & & \multicolumn{3}{c}{mAP} & & \multicolumn{3}{c}{mnDCG\textsubscript{1998}} \\
        \cmidrule(lr){3-5} \cmidrule(lr){7-9}
        Model & & Pixel & Blur & Mask & & Pixel & Blur & Mask \\
        \midrule
        \multicolumn{9}{c}{\textbf{With original query images}} \\
        \midrule
        Unadapted & & 51.7 & \textbf{71.6} & 37.5 & & 85.2 & \textbf{93.6} & 77.9 \\
        Adaption A & & 40.7 & 55.4 & 15.0 & & 80.8 & 88.1 & 58.8 \\
        Adaption B & & 36.4 & 44.4 & 22.9 & & 78.8 & 83.2 & 66.3 \\
        Adaption C & & 37.2 & 43.0 & 20.7 & & 79.2 & 82.5 & 64.2 \\
        \midrule
        \multicolumn{9}{c}{\textbf{With anonymized query images}} \\
        \midrule
        Unadapted & & 52.8 & \textbf{65.5} & 41.5 & & 86.4 & \textbf{91.5} & 81.0 \\
        Adaption A & & 43.8 & 55.0 & 22.1 & & 82.8 & 87.8 & 67.8 \\
        Adaption B & & 38.0 & 46.0 & 33.3 & & 79.8 & 84.1 & 76.9 \\
        Adaption C & & 40.4 & 45.1 & 32.1 & & 81.2 & 83.6 & 76.3 \\
        \bottomrule
    \end{tabular}
\end{table}
\begin{table}[ht]
    \centering
    \caption[RVL-CDIP document classification accuracy @ 50\% anonymization]{RVL-CDIP document classification accuracy @ 50\% anonymization.}
    \label{tab:rvl_cdip_document_classification_accuracy_at_50}
    \begin{tabular}{@{}llcccclccc@{}}
        \toprule
        & & \multicolumn{3}{c}{k-NN} & & \multicolumn{3}{c}{Linear} \\
        \cmidrule(lr){3-5} \cmidrule(lr){7-9}
        Model & & Pixel & Blur & Mask & & Pixel & Blur & Mask \\
        \midrule
        Unadapted & & 60.6 & 66.1 & 58.1 & & 58.1 & 61.7 & 57.0 \\
        Adaption A & & 59.2 & \textbf{66.8} & 38.6 & & 57.1 & \textbf{62.2} & 46.3 \\
        Adaption B & & 56.3 & 62.0 & 56.5 & & 55.5 & 58.7 & 56.6 \\
        Adaption C & & 58.2 & 60.9 & 56.0 & & 56.7 & 58.1 & 55.8 \\
        \bottomrule
    \end{tabular}
\end{table}
\begin{table}[ht]
    \centering
    \caption[RVL-CDIP retrieval qualities @ 75\% anonymization]{RVL-CDIP retrieval qualities @ 75\% anonymization.}
    \label{tab:rvl_cdip_retrieval_qualities_at_75}
    \begin{tabular}{@{}llcccclccc@{}}
        \toprule
        & & \multicolumn{3}{c}{mAP} & & \multicolumn{3}{c}{mnDCG\textsubscript{1998}} \\
        \cmidrule(lr){3-5} \cmidrule(lr){7-9}
        Model & & Pixel & Blur & Mask & & Pixel & Blur & Mask \\
        \midrule
        \multicolumn{9}{c}{\textbf{With original query images}} \\
        \midrule
        Unadapted & & 41.0 & \textbf{59.5} & 31.4 & & 79.4 & \textbf{89.4} & 73.8 \\
        Adaption A & & 36.2 & 49.3 & 7.0 & & 77.7 & 85.3 & 51.5 \\
        Adaption B & & 29.4 & 32.0 & 20.1 & & 72.9 & 75.5 & 63.9 \\
        Adaption C & & 27.4 & 29.4 & 17.6 & & 71.5 & 73.5 & 60.8 \\
        \midrule
        \multicolumn{9}{c}{\textbf{With anonymized query images}} \\
        \midrule
        Unadapted & & 43.8 & \textbf{56.9} & 37.0 & & 82.1 & \textbf{88.4} & 78.7 \\
        Adaption A & & 41.5 & 50.7 & 12.9 & & 81.5 & 86.0 & 60.6 \\
        Adaption B & & 34.9 & 38.1 & 31.1 & & 77.9 & 80.0 & 75.7 \\
        Adaption C & & 32.9 & 38.3 & 31.2 & & 76.7 & 80.2 & 75.8 \\
        \bottomrule
    \end{tabular}
\end{table}
\begin{table}[ht]
    \centering
    \caption[RVL-CDIP document classification accuracy @ 75\% anonymization]{RVL-CDIP document classification accuracy @ 75\% anonymization.}
    \label{tab:rvl_cdip_document_classification_accuracy_at_75}
    \begin{tabular}{@{}llcccclccc@{}}
        \toprule
        & & \multicolumn{3}{c}{k-NN} & & \multicolumn{3}{c}{Linear} \\
        \cmidrule(lr){3-5} \cmidrule(lr){7-9}
        Model & & Pixel & Blur & Mask & & Pixel & Blur & Mask \\
        \midrule
        Unadapted & & 57.0 & 65.2 & 56.8 & & 55.6 & 61.5 & 55.7 \\
        Adaption A & & 57.6 & \textbf{66.0} & 34.6 & & 55.7 & \textbf{62.2} & 41.7 \\
        Adaption B & & 54.9 & 58.5 & 57.0 & & 55.1 & 56.8 & 55.9 \\
        Adaption C & & 53.2 & 57.3 & 56.5 & & 54.0 & 56.6 & 55.5 \\
        \bottomrule
    \end{tabular}
\end{table}
\Figure[th!](topskip=0pt, botskip=0pt, midskip=0pt)[width=3.3in]{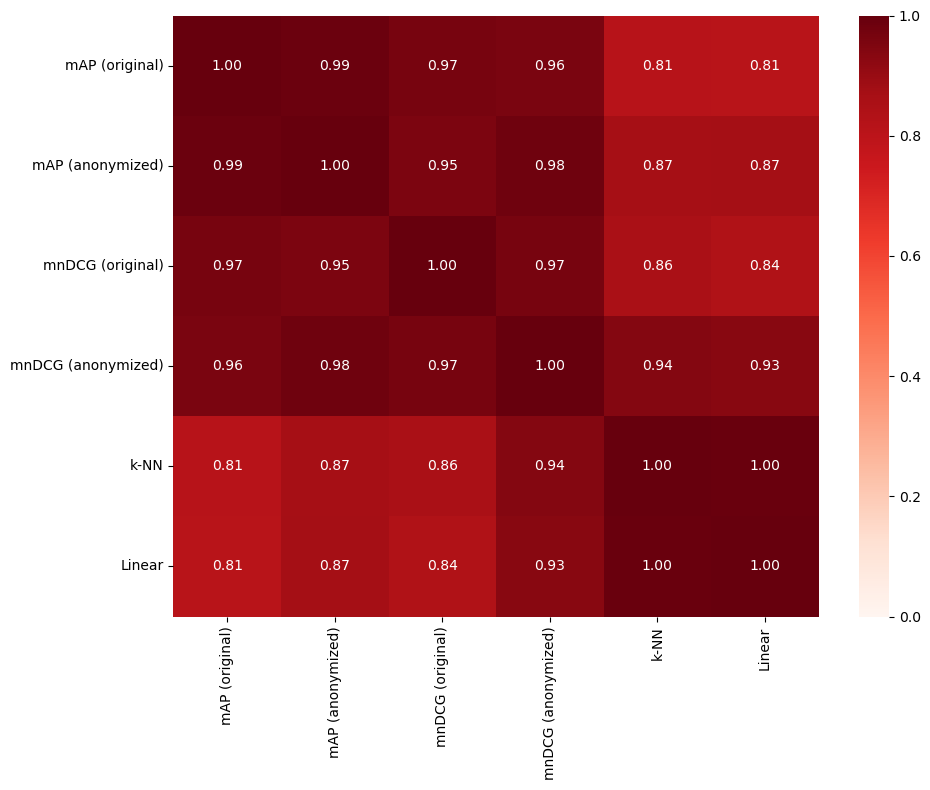}
{\textbf{RVL-CDIP metrics correlation matrix @ 50\% anonymization.}\label{fig:rvl_cdip_50_corr_matrix}}
\Figure[th!](topskip=0pt, botskip=0pt, midskip=0pt)[width=3.3in]{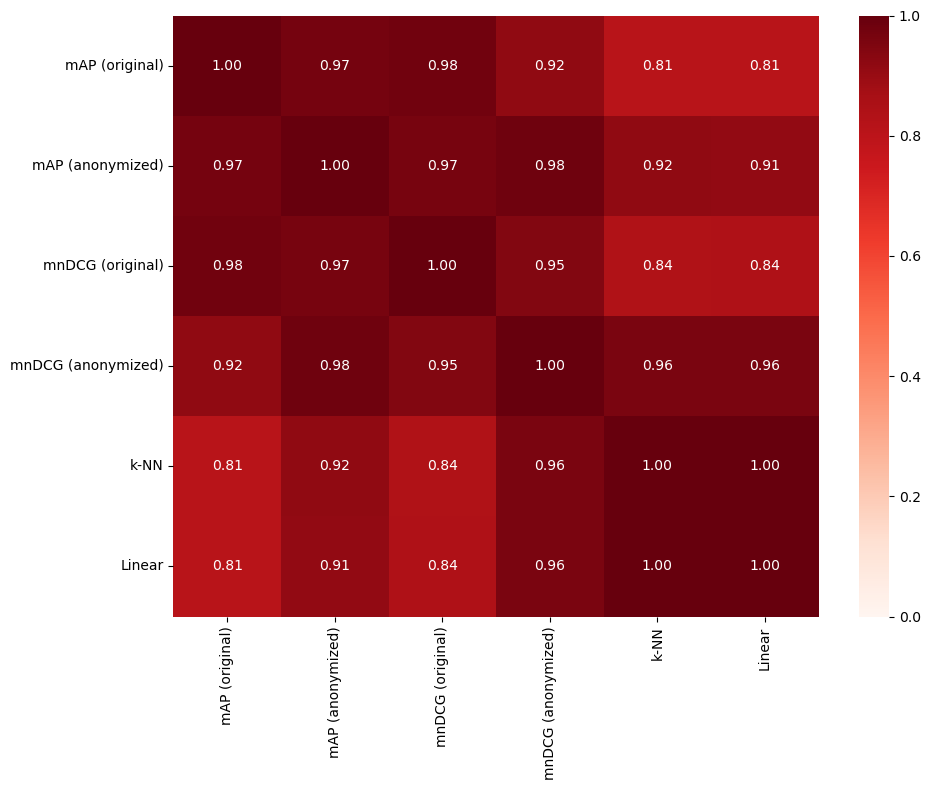}
{\textbf{RVL-CDIP metrics correlation matrix @ 75\% anonymization.}\label{fig:rvl_cdip_75_corr_matrix}}

\bibliographystyle{IEEEtran}
\bibliography{references}
\begin{IEEEbiography}[{\includegraphics[width=1in,height=1.25in,clip,keepaspectratio]{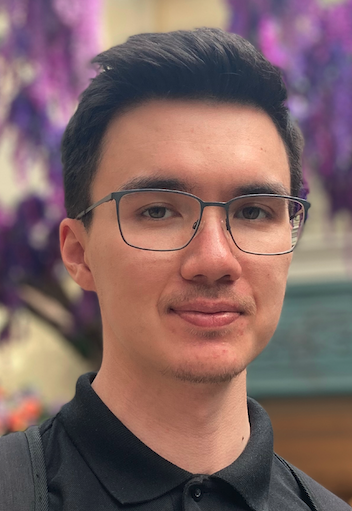}}]{Marvin Chen} received the M.Sc. degree in Computer Science and Media from Stuttgart Media University (HdM), Germany. He is currently working as a software developer at IBM Deutschland Research \& Development GmbH in Böblingen, Germany.

\end{IEEEbiography}
\begin{IEEEbiography}[{\includegraphics[width=1in,height=1.25in,clip,keepaspectratio]{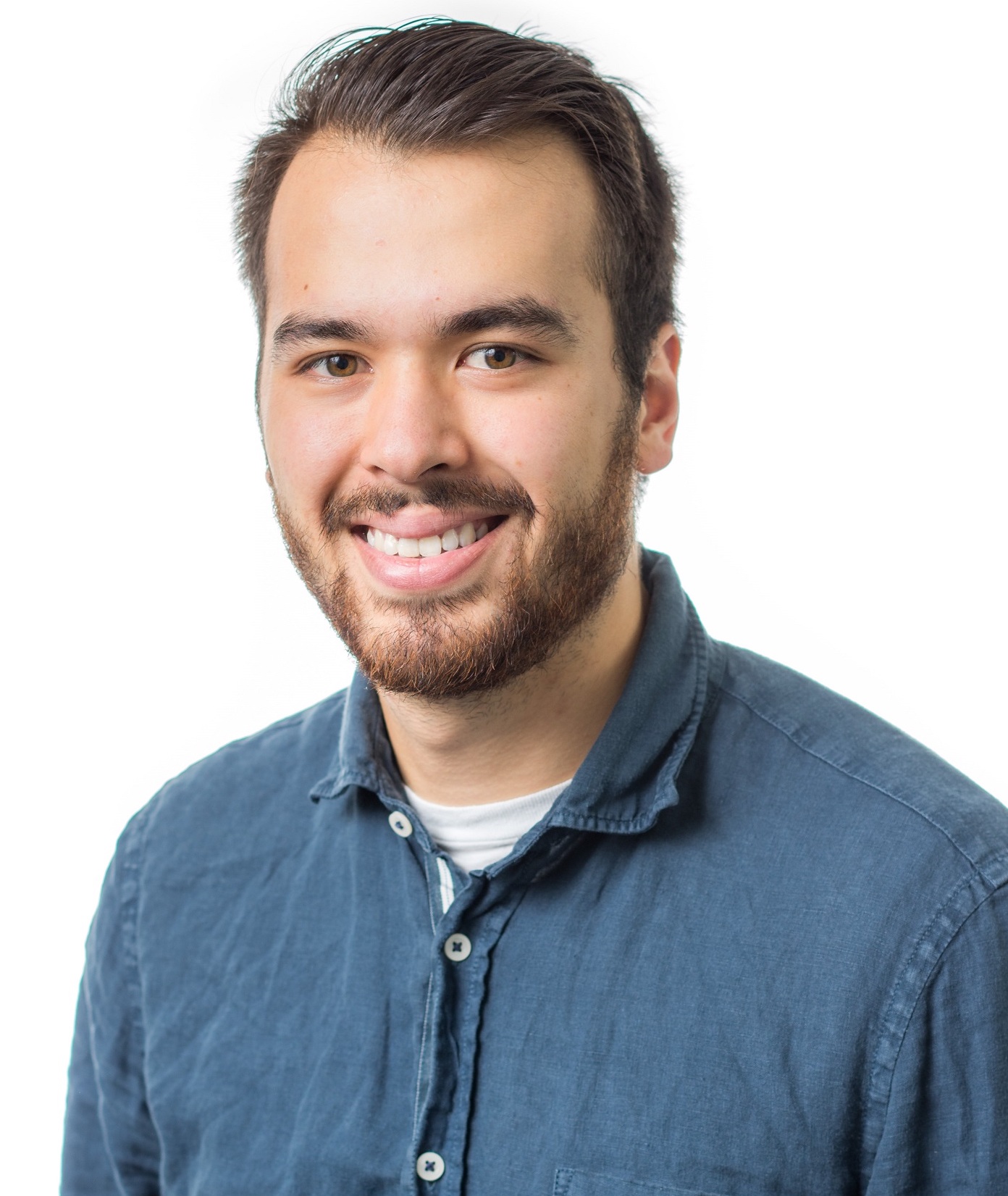}}]{Manuel Eberhardinger} received the M.Sc. degree in Data Science and Business Analytics  from Stuttgart Media University (HdM), Germany, where he is currently pursuing the Ph.D. degree in artificial intelligence (AI) with the Institute for Applied AI in combination with the Chair of Learning Technical Systems at the Ruhr University Bochum. His research focuses on explainable AI for sequential decision making and computer vision methods for document verification.

\end{IEEEbiography}
\begin{IEEEbiography}[{\includegraphics[width=1in,height=1.25in,clip,keepaspectratio]{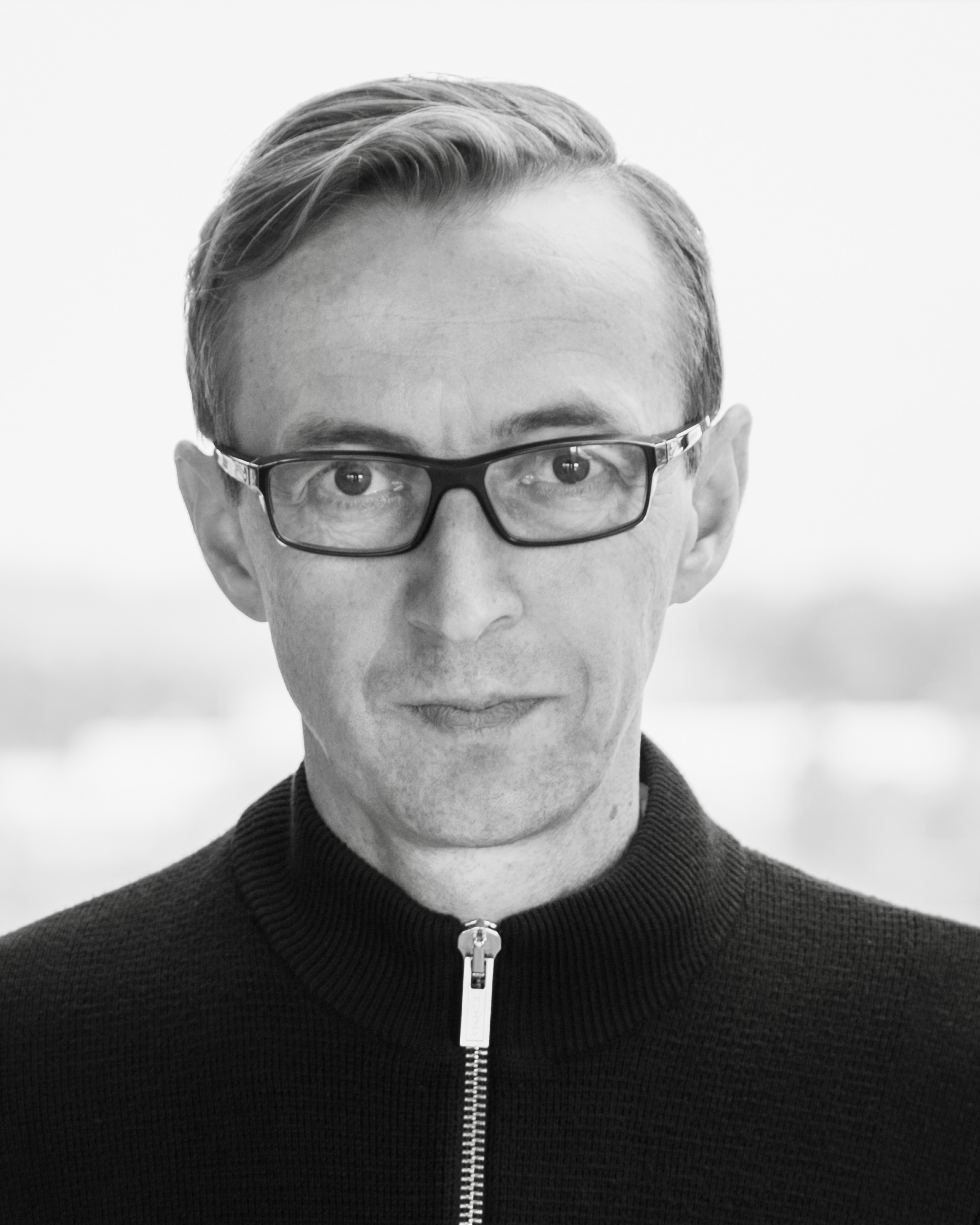}}]{Johannes Maucher} studied electrical engineering at the University of Ulm (Diploma). From the same university he received the Ph.D. degree at the Institute for Information Technology. After the Ph.D. he worked for 5 years at the Swiss technology group Ascom Systec AG, initially as a research engineer and later as a project manager and group leader. 
Since 2004 Johannes Maucher is a professor at Stuttgart Media University (HdM). There he is responsible for teaching and research activities in the field of artificial intelligence and machine learning on the Media Informatics (Bachelor) and Computer Science and Media (Master) degree programs. At HdM, he heads the Institute for Applied Artificial Intelligence (IAAI), which was founded in 2019. In his research work, Johannes Maucher investigates the application of deep neural networks for tasks in the fields of object recognition and natural language processing.

\end{IEEEbiography}
\EOD
\end{document}